\documentclass[runningheads]{llncs}
\usepackage{graphicx}
\usepackage{amsmath,amssymb} 
\usepackage{color}
\usepackage{multirow}
\usepackage{makecell}
\usepackage{booktabs}
\usepackage{colortbl}
\usepackage{subcaption}
\usepackage[pagebackref,breaklinks,colorlinks]{hyperref}
\usepackage{url}
\usepackage{orcidlink}
\definecolor{mygray}{RGB}{220, 220, 220}

\begin{document}
\title{SymmNeRF: Learning to Explore Symmetry Prior for Single-View View Synthesis}
\titlerunning{SymmNeRF}
%
\author{Xingyi Li\inst{1}\orcidlink{0000-0001-5765-3852} \and
Chaoyi Hong\inst{1}\orcidlink{0000-0003-1306-6634} \and
Yiran Wang\inst{1}\orcidlink{0000-0002-2785-9638} \and
Zhiguo Cao\inst{1}\orcidlink{0000-0002-9223-1863} \and
Ke Xian\inst{2}\thanks{Corresponding author}\orcidlink{0000-0002-0884-5126} \and
Guosheng Lin\inst{2}\orcidlink{0000-0002-0329-7458}
}
\authorrunning{Li et al.}
%
\institute{Key Laboratory of Image Processing and Intelligent Control, Ministry of Education \\ School of AIA, Huazhong University of Science and Technology, China\\
\email{\{xingyi\_li,cyhong,wangyiran,zgcao\}@hust.edu.cn}\\
\and
S-lab, Nanyang Technological University
\\
\email{\{ke.xian,gslin\}@ntu.edu.sg}
}
\maketitle              
\begin{abstract}
We study the problem of novel view synthesis of objects from a single image. Existing methods have demonstrated the potential in single-view view synthesis. However, they still fail to recover the fine appearance details, especially in self-occluded areas. This is because a single view only provides limited information. We observe that man-made objects usually exhibit symmetric appearances, which introduce additional prior knowledge. Motivated by this, we investigate the potential performance gains of explicitly embedding symmetry into the scene representation. In this paper, we propose SymmNeRF, a neural radiance field (NeRF) based framework that combines local and global conditioning under the introduction of symmetry priors.
In particular, SymmNeRF takes the pixel-aligned image features and the corresponding symmetric features as extra inputs to the NeRF, whose parameters are generated by a hypernetwork. As the parameters are conditioned on the image-encoded latent codes, SymmNeRF is thus scene-independent and can generalize to new scenes. Experiments on synthetic and real-world datasets show that SymmNeRF synthesizes novel views with more details regardless of the pose transformation, and demonstrates good generalization when applied to unseen objects. Code is available at: \href{https://github.com/xingyi-li/SymmNeRF}{\url{https://github.com/xingyi-li/SymmNeRF}}.

\end{abstract}

\section{Introdution}
Novel view synthesis is a long-standing problem in computer vision and graphics~\cite{debevec1996modeling,gortler1996lumigraph,levoy1996light}. The task is to synthesize novel views from a set of input views or even a single input view, which is challenging as it requires comprehensive 3D understanding~\cite{tewari2020state}. 
Prior works mainly focus on explicit geometric representations, such as voxel grids~\cite{jimenez2016unsupervised,liao2018deep,sitzmann2019deepvoxels,xie2019pix2vox}, point clouds~\cite{achlioptas2018learning,fan2017point}, and triangle meshes~\cite{kanazawa2018learning,ranjan2018generating,wang2018pixel2mesh}.
However, these methods suffer from limited spatial resolution and representation capability because of the discrete properties. Recently, differentiable neural rendering methods~\cite{mescheder2019occupancy,niemeyer2020differentiable,park2019deepsdf,saito2019pifu,sitzmann2019scene,Trevithick_2021_ICCV} have shown great progress in synthesizing photo-realistic novel views. For example, neural radiance fields (NeRFs)~\cite{mildenhall2020nerf}, which implicitly encode volumetric density and color via multi-layer perceptrons (MLPs), show an impressive level of fidelity on novel view synthesis. However, these methods usually require densely captured views as input and test-time optimization, leading to poor generalization across objects and scenes. To reduce the strong dependency on dense inputs and enable better generalization to unseen objects, in this paper, we explore novel view synthesis of object categories from only a single image.

\begin{figure}[t]
    \centering
    \includegraphics[scale=0.43]{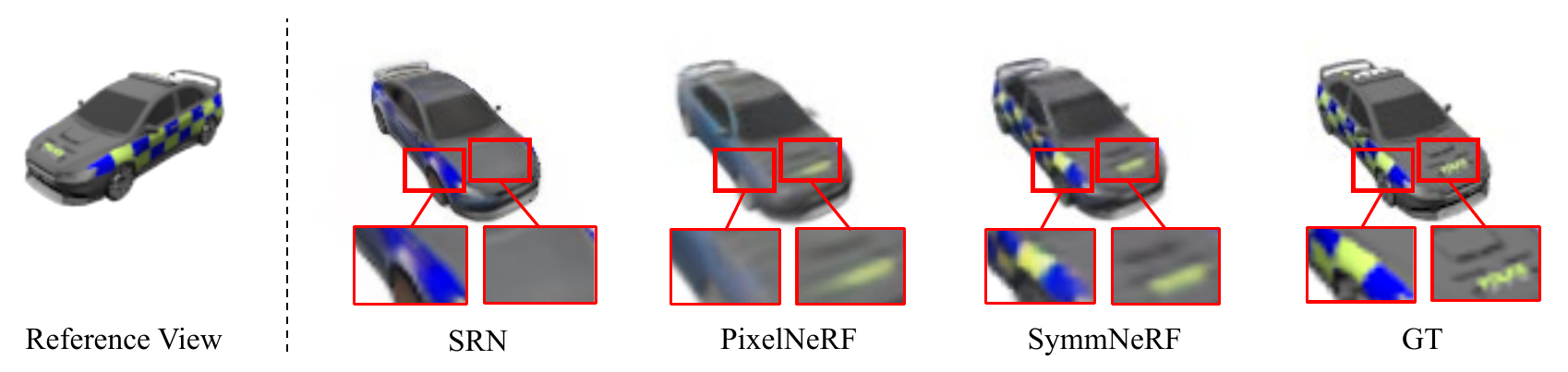}
    \caption{Novel views from a single image synthesized by SRN~\cite{sitzmann2019scene}, PixelNeRF~\cite{yu2021pixelnerf} and our SymmNeRF. The competitive methods are prone to miss some texture details, especially when the pose difference between the reference view and target view is large. By contrast, SymmNeRF augmented with symmetry priors recovers more appearance details.} 
    \label{fig:teaser}
\end{figure}

Novel view synthesis from a single image is challenging, because a single view cannot provide sufficient information. Recent NeRF-based methods~\cite{jang2021codenerf,yu2021pixelnerf} learn scene priors for reconstruction by training on multiple scenes. Although they have shown the potential in single-view view synthesis, it is particularly challenging when the pose difference between the reference and target view is large (see Fig.~\ref{fig:teaser}). We observe that man-made objects in real world usually exhibit symmetric appearances. Based on this, a question arises: {\textit {can symmetry priors benefit single-view view synthesis?}}

To answer this question, we explore how to take advantage of symmetry priors to introduce additional information for reconstruction. To this end, we present SymmNeRF, a NeRF-based framework that is augmented by symmetry priors. Specifically, we take the pixel-aligned image features and the corresponding symmetric features as extra inputs to NeRF. This allows reasonable recovery of occluded geometry and missing texture. During training, given a set of posed input images, SymmNeRF simultaneously optimizes a convolutional neural network (CNN) encoder and a hypernetwork. The former encodes image features, and generates latent codes which represent the coarse shape and appearance of unseen objects. The latter maps specific latent codes to the weights of the neural radiance fields. Therefore, SymmNeRF is scene-independent and can generalize to unseen objects. Unlike the original NeRF~\cite{mildenhall2020nerf}, for a single query point, we take as input its original and symmetric pixel-aligned image features besides its 3D location and viewing direction. At the inference stage, SymmNeRF generates novel views by feed-forward prediction without test-time optimization. 

In the present paper, we investigate the potential performance gains by combining local and global conditioning under the introduction of the symmetry prior. To this end, we add the assumptions on the data distribution that objects are in a canonical coordinate frame. 
We demonstrate that such a symmetry prior can lead to significant performance gains.
In summary, our main contributions are:
\begin{itemize}
    \item We propose SymmNeRF, a NeRF-based framework for novel view synthesis from a single image. By introducing symmetry priors into NeRF, SymmNeRF can synthesize high-quality novel views with fine details regardless of pose transformation.
    \item We combine local features with global conditioning via hypernetworks and demonstrate significant performance gains. Note that we perform inference via a CNN instead of auto-decoding, \textit{i.e.}, without test-time optimization, which is different from SRN~\cite{sitzmann2019scene}.
    \item Given only a single input image, SymmNeRF demonstrates significant improvement over state-of-the-art methods on synthetic and real-world datasets.
\end{itemize}

\section{Related Work}
\label{sec:related-work}
\paragraph{\rm {\textbf{Novel View Synthesis.}}}
Novel view synthesis is the task of synthesizing novel camera perspectives of a scene, given source images and their camera poses. The key challenges are understanding the 3D structure of the scene and inpainting of invisible regions of the scene~\cite{hani2020continuous}. 
The research of novel view synthesis has a long history in computer vision and graphics~\cite{debevec1996modeling,gortler1996lumigraph,levoy1996light}. Pioneer works typically synthesize novel views by warping, resampling, and blending reference views to target viewpoints, which can be classified as image-based rendering methods~\cite{debevec1996modeling}. However, they require densely captured views of the scene. When only a few observations are available, ghosting-like artifacts and holes may appear~\cite{tewari2020state}. With the advancement of deep learning, a few learning-based methods have been proposed, most of which focus on explicit geometric representations such as voxel grids~\cite{jimenez2016unsupervised,liao2018deep,sitzmann2019deepvoxels,xie2019pix2vox}, point clouds~\cite{achlioptas2018learning,fan2017point}, triangle meshes~\cite{kanazawa2018learning,ranjan2018generating,wang2018pixel2mesh}, and multiplane images (MPIs)~\cite{flynn2019deepview,tucker2020single,zhou2018stereo}. 

Recent works~\cite{chen2019learning,mescheder2019occupancy,niemeyer2020differentiable,park2019deepsdf,Trevithick_2021_ICCV} show that neural networks can be used as an implicit representation for 3D shapes and appearances. DeepSDF~\cite{park2019deepsdf} maps continuous spatial coordinates to signed distance and proves the superiority of neural implicit functions. 
SRN~\cite{sitzmann2019scene} proposes to represent 3D shapes and appearances implicitly as a continuous, differentiable function that maps a spatial coordinate to the local features of the scene properties at that point. 
Recently, NeRF~\cite{mildenhall2020nerf} shows astonishing results for novel view synthesis, which is an implicit MLP-based model that maps 3D coordinates plus 2D viewing directions to opacity and color values. However, NeRF requires enormous posed images and must be independently optimized for every scene. PixelNeRF~\cite{yu2021pixelnerf} tries to address this issue by conditioning NeRF on image features, which are extracted by an image encoder. This enables its ability to render novel views from only a single image and its generalization to new scenes. 
Rematas~\textit{et al.}~\cite{rematasICML21} propose ShaRF, a generative model aiming at estimating neural representation of objects from a single image, combining the benefits of explicit and implicit representations, which is capable of generalizing to unseen objects.
CodeNeRF~\cite{jang2021codenerf} learns to disentangle shape and texture by learning separate embeddings from a single image, allowing single view reconstruction and shape/texture editing. However, these methods usually struggle to synthesize reasonable novel views from a single image when self-occlusion occurs. In contrast, SymmNeRF first estimates coarse representations and then takes reflection symmetry as prior knowledge to inpaint invisible regions. This allows reasonable recovery of occluded geometry and missing texture.

\paragraph{\rm {\textbf{HyperNetworks.}}}
A hypernetwork~\cite{ha2016hypernetworks} refers to a small network that is trained to predict the weights of a large network, which has the potential to generalize previous knowledge and adapt quickly to novel environments. Various methods resort to hypernetworks in 3D vision. 
Littwin~\textit{et al.}~\cite{littwin2019deep} recover shape from a single image using hypernetworks in an end-to-end manner. SRN~\cite{sitzmann2019scene} utilizes hypernetworks for single-shot novel view synthesis with neural fields. In this work, we condition the parameters of NeRF on the image-encoded latent codes via the hypernetwork, which allows SymmNeRF to be scene-independent and generalize to new scenes.

\paragraph{\rm {\textbf{Reflection Symmetry.}}}
Reflection symmetry plays a significant role in the human visual system and has already been exploited in the computer vision community. 
Wu~\textit{et al.}~\cite{wu2020unsupervised} infer 3D deformable objects given only a single image, using a symmetric structure to disentangle depth, albedo, viewpoint and illumination.
Ladybird~\cite{xu2020ladybird} assigns occluded points with features from their symmetric points based on the reflective symmetry of the object, allowing recovery of occluded geometry and texture. NeRD~\cite{zhou2021nerd} learns a neural 3D reflection symmetry detector, which can estimate the normal vectors of objects’ mirror planes. 
They focus on the task of detecting the 3D reflection symmetry of a symmetric object from a 2D image. In this work, we focus on exploring the advantages of explicitly embedding symmetry into the scene representation for single-view view synthesis.

\section{SymmNeRF}
\label{subsec:method}

\begin{figure}[t]
    \centering
    \includegraphics[scale=0.49]{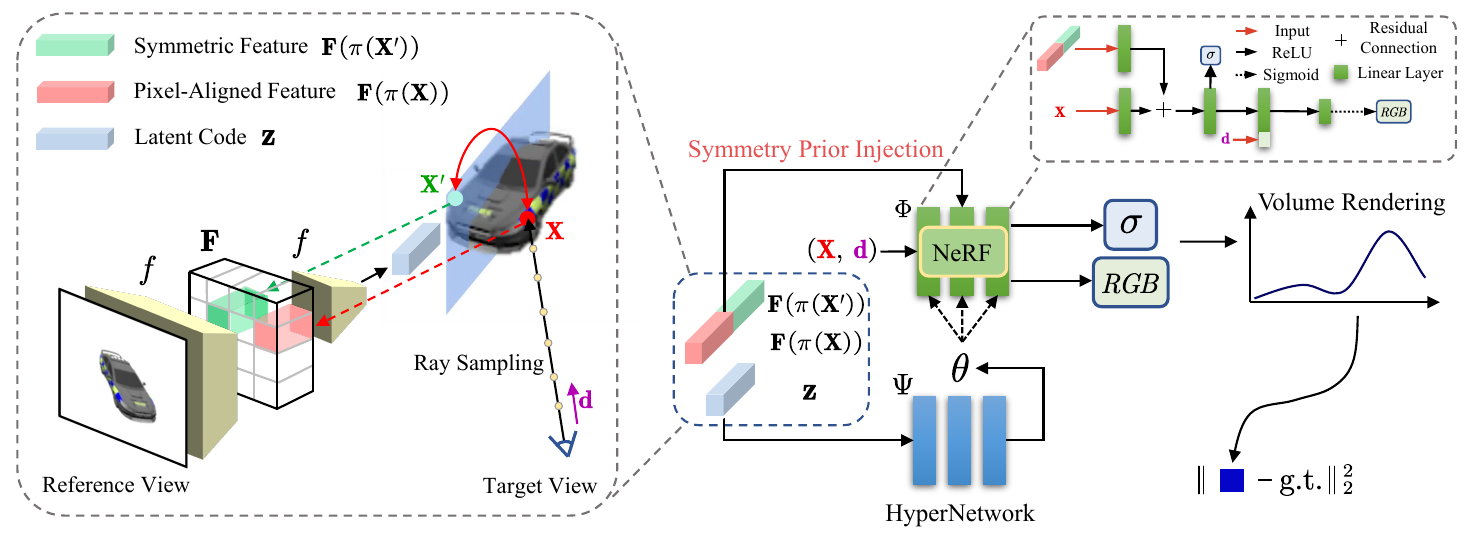}
    \caption{An overview of our SymmNeRF. Given a reference view, we first encode holistic representations by estimating the latent code $\mathbf{z}$ through the image encoder network $f$. We then obtain the pixel-aligned feature $\mathbf{F}(\pi(\mathbf{X}))$ and the symmetric feature $\mathbf{F}(\pi(\mathbf{X}^{\prime}))$, by projecting the query point $\mathbf{X}$ and the symmetric point $\mathbf{X}^{\prime}$ to the 2D location on the image plane using camera parameters, followed by bilinear interpolation between the pixelwise features on the feature volume $\mathbf{F}$. The hypernetwork transforms the latent code $\mathbf{z}$ to the weights $\theta$ of the corresponding NeRF $\rm \Phi$. For a query point $\mathbf{X}$ along a target camera ray with viewing direction $\mathbf{d}$, NeRF takes the spatial coordinate $\mathbf{X}$, ray direction $\mathbf{d}$, pixel-aligned feature $\mathbf{F}(\pi(\mathbf{X}))$ and symmetric feature $\mathbf{F}(\pi(\mathbf{X}^{\prime}))$ as input, and outputs the color and density.}
    \label{fig:pipeline}
\end{figure}

\subsection{Overview}
Here we present an overview of our proposed method. We propose to firstly estimate holistic representations as well as symmetry planes, 
followed by fulfilling details, and to explicitly inject symmetry priors into single-view view synthesis of object categories. In particular, we design SymmNeRF to implement the ideas above. Fig.~\ref{fig:pipeline} shows the technical pipeline of SymmNeRF.

Given a set of $M$ instances training datasets $\mathcal{D} = \{ \mathcal{C}_{j} \}_{j=1}^{M}$, 
where $\mathcal{C}_{j} = \{ (\mathcal{I}_{i}^{j}, \mathbf{E}_{i}^{j}, \mathbf{K}_{i}^{j}) \}_{i=1}^{N}$ is a dataset of an instance object, $\mathcal{I}_{i}^{j} \in \mathbb{R}^{H \times W \times 3}$ refers to an input image, $\mathbf{E}_{i}^{j} = [\mathbf{R} | \mathbf{t} ] \in \mathbb{R}^{3 \times 4}$ and $\mathbf{K}_{i}^{j} \in \mathbb{R}^{3 \times 3}$ are the corresponding extrinsic and intrinsic camera matrix respectively, and $N$ denotes the number of the input images, 
SymmNeRF first encodes a holistic representation and regresses the symmetry plane for the input view. 
We then extract symmetric features along with pixel-aligned features for the sake of preserving fine-grained details observed in the input view. Subsequently, we transform the holistic representation to the weights of the corresponding NeRF, and inject symmetry priors (\textit{i.e.}, symmetric features as well as pixel-aligned features) to predict colors and densities. Finally, we adopt the classic volume rendering technique~\cite{kajiya1984ray,mildenhall2020nerf} to synthesize novel views.

\subsection{Encoding Holistic Representations}
\label{sec:encoder}
Humans usually 
understand the 3D shapes and appearances by first generating a profile, then restoring details from observations. Emulating the human visual system, we implement coarse depictions of objects by an image encoder network. 

The image encoder network $f$ is responsible for mapping the input image $\mathcal{I}_{i}$ into the latent code $\mathbf{z}_{i}$ which characterizes holistic information of the object's shape and appearance:
\begin{equation}
    f: \mathbb{R}^{H \times W \times 3} \rightarrow \mathbb{R}^{k},\quad \mathcal{I}_{i} \mapsto f(\mathcal{I}_{i}) = \mathbf{z}_{i},
\end{equation}
where $k$ is the dimension of $\mathbf{z}_{i}$, and the parameters of $f$ are denoted by $\rm \Omega$. Here, we denote the feature volume extracted by $f$ during the encoding of holistic representations as $\mathbf{F}$ (\textit{i.e.}, the concatenation of upsampled feature maps outputted by ResNet blocks). The image encoder network contains four ResNet blocks of ResNet-34~\cite{he2016deep}, followed by an average pooling layer and a fully connected layer.

\subsection{Extracting Symmetric Features}
\label{sec:symmetry}
The holistic representations introduced in the previous section coarsely describe objects. To synthesize detailed novel views, we follow PixelNeRF~\cite{yu2021pixelnerf} and adopt pixel-aligned image features to compensate for the details. However, simply using pixel-aligned image features ignores the underlying 3D structure. In contrast, humans can infer the 3D shape and appearance from a single image, despite the information loss and self-occlusion that occurs during the imagery capture. This can boil down to the fact that humans usually resort to prior knowledge, 
\textit{e.g.}, symmetry.
Motivated by the above observation, we propose to inpaint invisible regions and alleviate the ill-posedness of novel view synthesis from a single image via symmetry priors. In the following, we briefly introduce the properties of 3D reflection symmetry~\cite{zhou2021nerd}, followed by how symmetry priors are applied.

\paragraph{\rm {\textbf{3D reflection symmetry.}}}
When two points on an object's surface are symmetric, they share identical surface properties of the object. Formally, we define the symmetry regrading a rigid transformation $\mathbf{M} \in \mathbb{R}^{4 \times 4}$ as
\begin{equation}
    \forall \mathbf{X} \in \mathbb{S}: 
        \begin{cases}
        \mathbf{MX} \in \mathbb{S}, \\
        \mathcal{F}(\mathbf{X}) = \mathcal{F}(\mathbf{MX}),
        \end{cases}
\end{equation}
where $\mathbf{X}$ is the homogeneous coordinate of a point on the object's surface, $ \mathbb{S} \subset \mathbb{R}^{4} $ is the set of points that are on the surface of an object, $\mathbf{MX}$ is the symmetric point of $\mathbf{X}$, and $\mathcal{F}(\cdot)$ stands for the surface properties at a given point.

The 2D projections $\mathbf{x} = [x,y,1,1/d]^T$ and $\mathbf{x}^{\prime} = [x^{\prime},y^{\prime},1,1/d^{\prime}]^T$ of two 3D points $\mathbf{X}, \mathbf{X}^{\prime} \in \mathbb{S}$ satisfy
\begin{equation}
    \begin{cases}
    \mathbf{x} = \mathbf{K} \mathbf{R_{t}} \mathbf{X} / d, \\
    \mathbf{x}^{\prime} = \mathbf{K} \mathbf{R_{t}} \mathbf{X}^{\prime} / d^{\prime},
    \end{cases}
\end{equation}
where $\mathbf{K} \in \mathbb{R}^{4 \times 4}$ and $\mathbf{R_{t}} \in \mathbb{R}^{4 \times 4}$ are respectively the camera intrinsic matrix and extrinsic matrix. The latter transforms the coordinate from the world coordinate system to the camera coordinate system. $d, d^{\prime}$ are the depth in the camera space. When these two points are symmetric w.r.t. a rigid transformation, \textit{i.e.}, $\mathbf{X}^{\prime} = \mathbf{MX}$, the following constraint can be derived:
\begin{equation}
    \mathbf{x}^{\prime} = \frac{d}{d^{\prime}}\mathbf{K} \mathbf{R_{t}} \mathbf{M} \mathbf{R_{t}}^{-1} \mathbf{K}^{-1} \mathbf{x},
\end{equation}
where $\mathbf{x}$ and $\mathbf{x}^{\prime}$ are 2D projections of these two 3D points. This suggests that given a 2D projection $\mathbf{x}$, we can obtain its symmetric counterpart $\mathbf{x}^{\prime}$ if $\mathbf{M}$ and camera parameters are known. In this paper, we focus on exploring the benefits of explicitly embedding symmetry into our representation. 
To this end, we add the assumptions on the data distribution that objects are in a canonical coordinate frame, and that their symmetry axis is known.

\paragraph{\rm {\textbf{Applying Symmetry Prior.}}}
To inpaint invisible regions, we apply the symmetry property introduced above and extract symmetric features $\mathbf{F}(\pi(\mathbf{X}^{\prime}))$. This can be achieved by projecting the symmetric point $\mathbf{X}^{\prime}$ to the 2D location $\mathbf{x}^{\prime}$ on the image plane using camera parameters, followed by bilinearly interpolating between the pixelwise features on the feature volume $\mathbf{F}$ extracted by the image encoder network $f$. In addition, we follow PixelNeRF~\cite{yu2021pixelnerf} and adopt pixel-aligned features $\mathbf{F}(\pi(\mathbf{X}))$, which share the same acquisition approach with symmetric features. They are subsequently concatenated together to form the local image features corresponding to $\mathbf{X}$.

\subsection{Injecting Symmetry Prior into NeRF}
In this section, we inject symmetry priors into the neural radiance field for single-view view synthesis. Technically, the weights of the neural radiance field are conditioned on the latent code $\mathbf{z}_{i}$ introduced in Sec.~\ref{sec:encoder}, which represents a coarse but holistic depiction of the object. To preserve fine-grained details, during the color and density prediction, we also take the pixel-aligned image features and the corresponding symmetric features as extra inputs to fulfill details observed in the input view.

\paragraph{\rm {\textbf{Generating Neural Radiance Fields.}}}
We generate a specific neural radiance field by mapping a latent code
$\mathbf{z}_{i}$ to the weights $\theta_{i}$ of the neural radiance field using the hypernetwork $\rm \Psi$, which can be defined as follows:
\begin{equation}
    {\rm \Psi}: \mathbb{R}^{k} \rightarrow \mathbb{R}^{l},\quad \mathbf{z}_{i} \mapsto {\rm \Psi}(\mathbf{z}_{i}) = \theta_{i},
\end{equation}
where, $l$ stands for the dimension of the parameter space of neural radiance fields. We parameterize $\rm \Psi$ as an MLP with parameters $\psi$. This can be interpreted as a simulation of the human visual system. Specifically, humans first estimate the holistic shape and appearance of the unseen object when given a single image, then formulate a sketch in their mind to represent the object. Similarly, SymmNeRF encodes overall information of the object as a latent code from a single image, followed by generating a corresponding neural radiance field to describe the object.

\paragraph{\rm {\textbf{Color and Density Prediction.}}}
Given a reference image with known camera parameters, for a single query point location $\mathbf{X} \in \mathbb{R}^{3}$ on a ray $\mathbf{r} \in \mathbb{R}^{3}$ with unit-length viewing direction $\mathbf{d} \in \mathbb{R}^{3}$, SymmNeRF predicts the color and density at that point in 3D space, which is defined as:
\begin{equation}
    \begin{gathered}
        {\rm \Phi}: \mathbb{R}^{m_{\mathbf{X}}} \times \mathbb{R}^{m_{\mathbf{d}}} \times \mathbb{R}^{2n} \rightarrow \mathbb{R}^{3} \times \mathbb{R}, \\
        (\gamma_{\mathbf{X}}(\mathbf{X}), \gamma_{\mathbf{d}}(\mathbf{d}), \mathbf{F}(\pi(\mathbf{X})), \mathbf{F}(\pi(\mathbf{X}^{\prime}))) \mapsto \\
        {\rm \Phi}(\gamma_{\mathbf{X}}(\mathbf{X}), \gamma_{\mathbf{d}}(\mathbf{d}), \mathbf{F}(\pi(\mathbf{X})), \mathbf{F}(\pi(\mathbf{X}^{\prime}))) = (\mathbf{c}, \sigma),
    \end{gathered}
\end{equation}
where $\rm \Phi$ represents a neural radiance field, an MLP network whose weights are given by the hypernetwork $\rm \Psi$, $\mathbf{X}^{\prime} \in \mathbb{R}^{3}$ is the corresponding symmetric 3D point of $\mathbf{X}$, $\gamma_{\mathbf{X}}(\cdot)$ and $\gamma_{\mathbf{d}}(\cdot)$ are positional encoding functions for spatial locations and viewing directions, $n$, ${m_{\mathbf{X}}}$ and ${m_{\mathbf{d}}}$ are respectively the dimensions of pixel-aligned features (symmetric features), $\gamma_{\mathbf{X}}(\mathbf{X})$ and $\gamma_{\mathbf{d}}(\mathbf{d})$, $\pi$ denotes the process of projecting the 3D point onto the image plane using known intrinsics, and $\mathbf{F}$ is the feature volume extracted by the image encoder network $f$. 

\subsection{Volume Rendering}
To render the color of a ray $\mathbf{r}$ passing through the scene, we first compute its camera ray $\mathbf{r}$ using the camera parameters and sample $K$ points $\{\mathbf{X}_{k}\}_{k=1}^{K}$ along the camera ray $\mathbf{r}$ between near and far bounds, and then perform classical volume rendering~\cite{kajiya1984ray,mildenhall2020nerf}:
\begin{gather}
    \Tilde{C}(\mathbf{r}) = \sum_{k=1}^{K} T_{k}(1 - \text{exp}(-\sigma_{k} \delta_{k})) \mathbf{c}_{k}, \\
    \text{where} \quad T_{k} = \text{exp}(-\sum_{j=1}^{k-1}\sigma_{j} \delta_{j}),
\end{gather}
where $\mathbf{c}_{k}$ and $\sigma_{k}$ denote the color and density of the $k$-th sample on the ray, respectively, and $\delta_{k} = \Vert \mathbf{X}_{k+1} - \mathbf{X}_{k} \Vert_2$ is the interval between adjacent samples.

\subsection{Training}
To summarize, given a set of $M$ instances training datasets $\mathcal{D} = \{ \mathcal{C}_{j} \}_{j=1}^{M}$, where $\mathcal{C}_{j} = \{ (\mathcal{I}_{i}^{j}, \mathbf{E}_{i}^{j}, \mathbf{K}_{i}^{j}) \}_{i=1}^{N}$ is a dataset of an instance object, we optimize SymmNeRF to minimize the rendering error of observed images:
\begin{gather}
    \min_{{\rm \Omega}, \psi} \sum_{j=1}^{M} \sum_{i=1}^{N} \mathcal{L}(\mathcal{I}_{i}^{j}, \mathbf{E}_{i}^{j}, \mathbf{K}_{i}^{j}; {\rm \Omega}, \psi), \\
    \mathcal{L} = \sum_{\mathbf{r} \in \mathcal{R}} \Big\Vert \Tilde{C}(\mathbf{r}) - C(\mathbf{r}) \Big\Vert_2^{2},
\end{gather}
where $\rm \Omega$ and $\psi$ are respectively the parameters of the image encoder network $f$ and the hypernetwork $\rm \Psi$, $\mathcal{R}$ is the set of camera rays passing through image pixels, and $C(\mathbf{r})$ denotes the ground truth pixel color.

\section{Experiments}
\label{sec:experiments}
\subsection{Datasets}
\paragraph{\rm {\textbf{Synthetic Renderings.}}}
We evaluate our approach on the synthetic ShapeNet benchmark~\cite{chang2015shapenet} for single-shot reconstruction. 1) We mainly focus on the ShapeNet-SRN dataset, following the same protocol adopted in~\cite{sitzmann2019scene}. 
This dataset includes two object categories: $3,514$ ``Cars'' and $6,591$ ``Chairs''. The train/test split is predefined across object instances. There are $50$ views per object instance in the training set. For testing, $251$ novel views in an Archimedean spiral are used for each object instance in the test set. All images are at $128 \times 128$ pixels; 2) Similar to PixelNeRF~\cite{yu2021pixelnerf}, we also test our method on the ShapeNet-NMR dataset~\cite{kato2018renderer} under two settings: category-agnostic single-view reconstruction and generalization to unseen categories, 
following~\cite{kato2018renderer,liu2019softras,niemeyer2020differentiable}. 
This dataset contains the $13$ largest categories of ShapeNet and provides $24$ fixed elevation views for each object instance. All images are of $64 \times 64$ resolution.

\paragraph{\rm {\textbf{Real-World Renderings.}}}
We also generalize our model, trained only on the ShapeNet-SRN dataset, directly to two complex real-world datasets. One is the Pix3D~\cite{sun2018pix3d} dataset containing various image-shape pairs with 2D-3D alignment. The other is the Stanford Cars~\cite{krause20133d} dataset which contains various real images of $196$ classes of cars. All images of the two datasets are cropped and resized to $128 \times 128$ pixels during testing.

\subsection{Implementation Details}
SymmNeRF is trained using the AdamW optimizer~\cite{kingma2014adam,loshchilov2018decoupled}. The learning rate follows the warmup~\cite{he2016deep} strategy: linearly growing from $0$ to $1 \times 10^{-4}$ during the first $2$k iterations and then decaying exponentially close to $0$ over the optimization. The network parameters are updated with around $400$-$500$k iterations. We use a batch size of $4$ objects and a ray batch size of $256$, each queried at $64$ samples. Experiments are conducted on $2$ NVIDIA GeForce RTX $3090$ GPUs.

\begin{table}[t]
\footnotesize
\centering 
	\caption{Quantitative comparisons against state-of-the-art methods on ``Cars'' and ``Chairs'' classes of the ShapeNet-SRN dataset. 
	The best performance is in \textbf{bold}, and the second best is \underline{underlined}.}
	\resizebox{0.62\linewidth}{!}{
		\renewcommand\arraystretch{0.85}	
		\begin{tabular}{ccccccc}
			\toprule
			& \multicolumn{2}{c}{Chairs} & \multicolumn{2}{c}{Cars} & \multicolumn{2}{c}{Average}\\
			\cmidrule{2-7}
			Methods & PSNR$\uparrow$ & SSIM$\uparrow$ & PSNR$\uparrow$ & SSIM$\uparrow$ & PSNR$\uparrow$ & SSIM$\uparrow$ \\
			\midrule

			\cellcolor{mygray}GRF~\cite{Trevithick_2021_ICCV} (ICCV'21) & \cellcolor{mygray}21.25 & \cellcolor{mygray}0.86  & \cellcolor{mygray}20.33 & \cellcolor{mygray}0.82 & \cellcolor{mygray}20.79 & \cellcolor{mygray}0.84 \\
			
			TCO~\cite{tatarchenko2016multi} (ECCV'16) & 21.27 & 0.88 & - & - & - & -\\

			\cellcolor{mygray}dGQN~\cite{eslami2018neural} (Science'18) & \cellcolor{mygray}21.59 & \cellcolor{mygray}0.87 & \cellcolor{mygray}- & \cellcolor{mygray}- & \cellcolor{mygray}- & \cellcolor{mygray}- \\
			
			ENR~\cite{dupont2020equivariant} (ICML'20) & 22.83 & - & 22.26 & - & 22.55 & - \\

			\cellcolor{mygray}SRN~\cite{sitzmann2019scene} (NeurIPS'19) & \cellcolor{mygray}22.89 &  \cellcolor{mygray}0.89 & \cellcolor{mygray}22.25 & \cellcolor{mygray}0.89 & \cellcolor{mygray}22.57 & \cellcolor{mygray}0.89 \\
			
			PixelNeRF~\cite{yu2021pixelnerf} (CVPR'21) & \underline{23.72} & \underline{0.91} & 23.17 & \underline{0.90} & 23.45 & \underline{0.91} \\

			\cellcolor{mygray}ShaRF~\cite{rematasICML21} (ICML'21) & \cellcolor{mygray}23.37 & \cellcolor{mygray}\textbf{0.92} & \cellcolor{mygray}22.53 & \cellcolor{mygray}\underline{0.90} & \cellcolor{mygray}22.90 & \cellcolor{mygray}\underline{0.91}\\
			
			CodeNeRF~\cite{jang2021codenerf} (ICCV'21) & 23.66 & 0.90 & \textbf{23.80} & \textbf{0.91} & \underline{23.74} & \underline{0.91}\\
			
			\cellcolor{mygray}Ours & \cellcolor{mygray}\textbf{24.32} & \cellcolor{mygray}\textbf{0.92} & \cellcolor{mygray}\underline{23.44} & \cellcolor{mygray}\textbf{0.91} & \cellcolor{mygray}\textbf{23.88} & \cellcolor{mygray}\textbf{0.92} \\
			
			\bottomrule
		\end{tabular}
	}
	\label{tab:quantitative-shapenet-srn}
\end{table}

\begin{figure}
    \begin{minipage}[t]{.47\linewidth}
        \centering
        \includegraphics[scale=0.08]{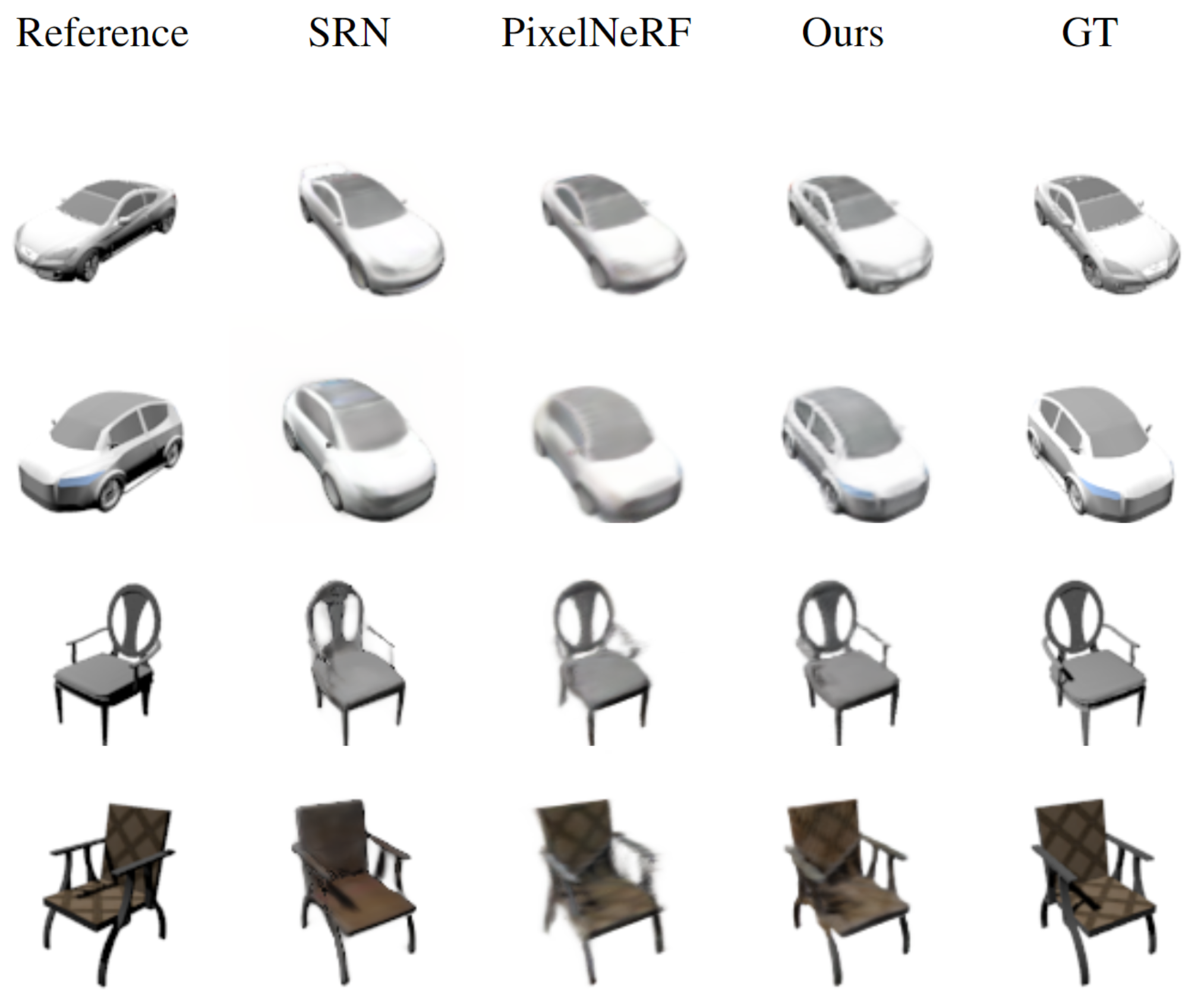}
        \caption{Qualitative comparisons on ``Cars'' and ``Chairs'' classes. SymmNeRF can produce high-quality renderings with fine-grained details, proper geometry and reasonable appearance.}
        \label{fig:qualititive}
    \end{minipage}~~
    \begin{minipage}[t]{.5\linewidth}
        \centering
        \includegraphics[scale=0.11]{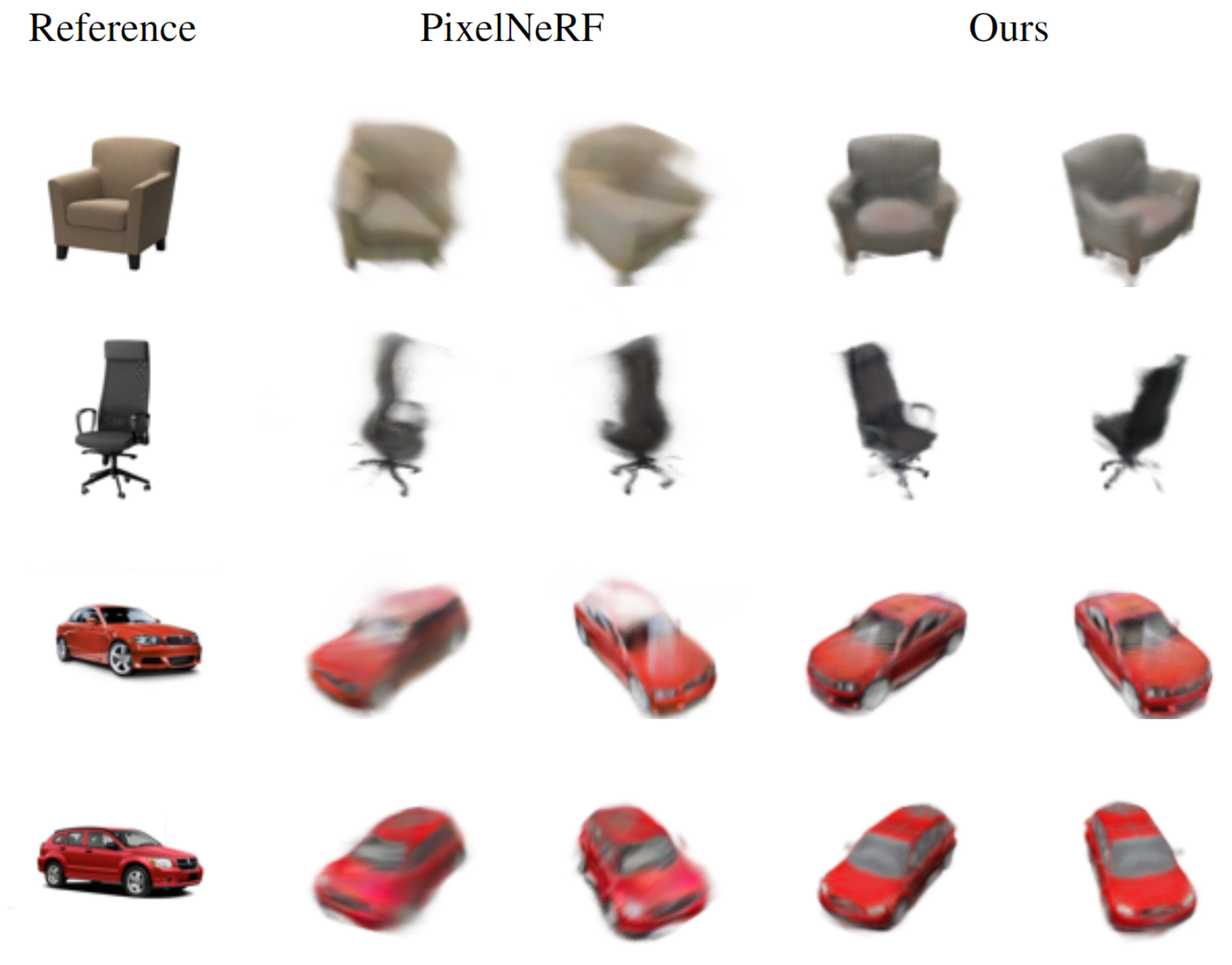}
        \caption{Qualitative comparisons with PixelNeRF~\cite{yu2021pixelnerf} on real-world Pix3D~\cite{sun2018pix3d} and Stanford Cars~\cite{krause20133d} datasets. Compared with PixelNeRF, SymmNeRF yields better generalization.}
        \label{fig:real-world}
    \end{minipage}
\end{figure}

\subsection{Comparisions}
Here we compare SymmNeRF against the existing state-of-the-art methods, among which CodeNeRF, SRN and ShaRF require test-time optimization at inference, and ShaRF entails 3D ground truth voxel grids besides 2D supervision. 
To evaluate the quality of renderings, we adopt two standard image quality metrics: the Peak Signal-to-Noise Ratio (PSNR) and the Structure Similarity Index Measure (SSIM)~\cite{wang2004image}. We also include LPIPS~\cite{zhang2018perceptual} in all experiments except the ShapeNet-SRN dataset. The better approach favors the higher PSNR and SSIM, and the lower LPIPS. Please refer to the supplementary material for more visualization.

\paragraph{\rm {\textbf{Evaluations on the ShapeNet-SRN Dataset.}}}
In general, as shown in Table~\ref{tab:quantitative-shapenet-srn}, our method outperforms or at least is on par with state-of-the-art methods. For the ``Cars'' category, SymmNeRF outperforms its competitors including PixelNeRF, SRN and ShaRF, and achieves comparable performance with CodeNeRF. Note that \textit{our SymmNeRF solves a much harder problem than SRN and CodeNeRF}. In particular, SymmNeRF directly infers the unseen object representation in a single forward pass, while SRN and CodeNeRF need to be retrained on all new objects to optimize the latent codes. In addition, we observe that most cars from the ``Cars'' category share similar 3D shapes and simple textures. As a result, the experiment on the ``Cars'' category is in favor of CodeNeRF. In contrast, for the ``Chairs'' category, SymmNeRF significantly outperforms all baselines across all metrics by a large margin. This result implies that our model can generalize well on new objects, as the shapes and textures of chairs in the ``Chairs'' category vary considerably. 
This implies that SymmNeRF indeed captures the underlying 3D structure of objects with the help of symmetry priors and the hypernetwork, rather than simply exploiting data biases.

Here we compare SymmNeRF qualitatively with SRN and PixelNeRF in Fig.~\ref{fig:qualititive}. One can observe that: i) SRN is prone to generate overly smooth renderings and is unable to capture the accurate geometry, leading to some distortions; ii) PixelNeRF performs well when the query view is close to the reference one, but fails to recover the details invisible in the reference, especially when the rendered view is far from the reference; iii) SymmNeRF, by contrast, can synthesize photo-realistic, reasonable novel views with fine-grained details close to ground truths.

\begin{figure*}[t]
    \centering
    \includegraphics[scale=0.13]{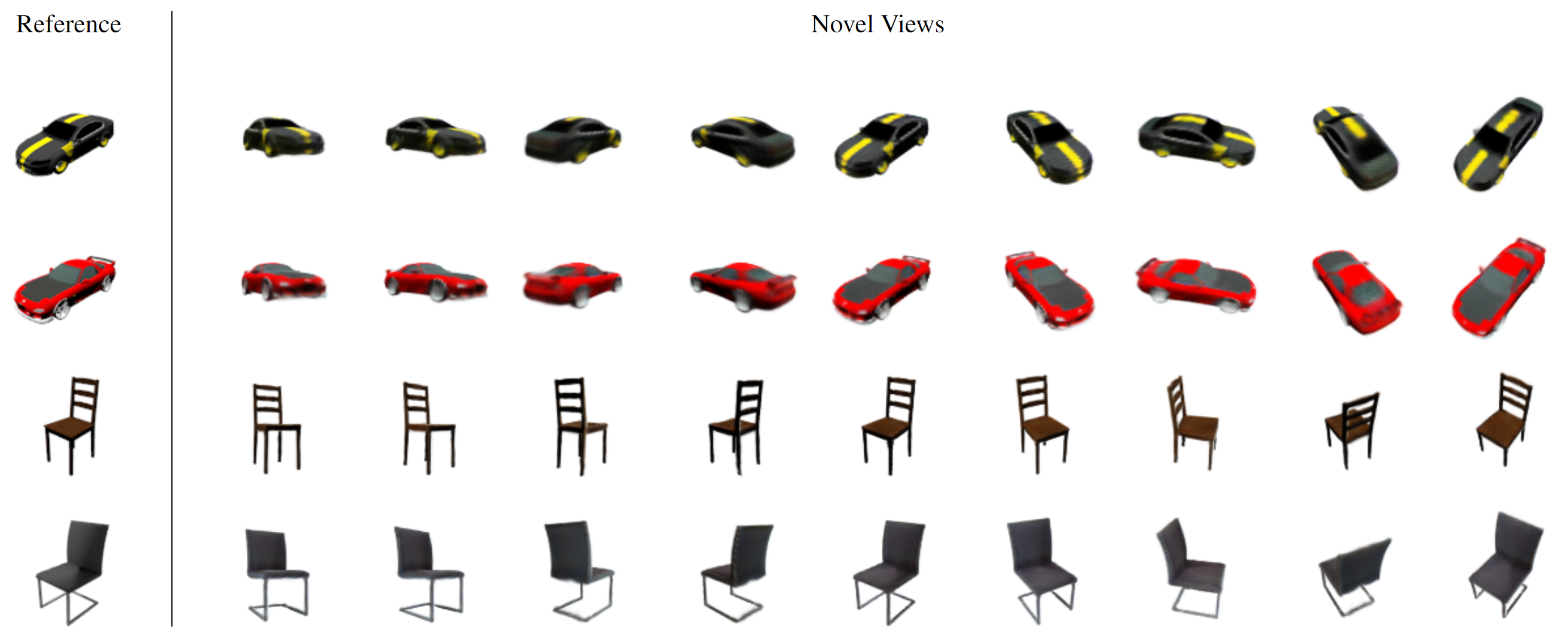}
    \caption{Novel view synthesis on ``Cars'' and ``Chairs'' of ShapeNet-SRN dataset.}
    \label{fig:novel-view-synthesis}
\end{figure*}

We further demonstrate the high-quality results of SymmNeRF by providing more novel view synthesis visualization in Fig.~\ref{fig:novel-view-synthesis}. As can be seen, SymmNeRF can always synthesize photo-realistic and reasonable novel renderings from totally different viewpoints. 

\paragraph{\rm {\textbf{Generalization on Real-World Datasets.}}}
To further investigate the generalization of SymmNeRF, we evaluate SymmNeRF on two real-world datasets, \textit{i.e.}, the Pix3D~\cite{sun2018pix3d} and the Stanford Cars~\cite{krause20133d}. Note that for the lack of ground truth, we only show the qualitative results on the two datasets.
Here we apply SymmNeRF trained on the synthetic chairs and cars directly on the real-world images without any finetuning.
As shown in Fig.~\ref{fig:real-world}, PixelNeRF~\cite{yu2021pixelnerf} fails to synthesize reasonable novel views, because it only notices the use of pixel-aligned image features, ignoring the underlying 3D structure the reference view provides. Compared with PixelNeRF~\cite{yu2021pixelnerf}, SymmNeRF can effectively infer the geometry and appearance of real-world chairs and cars. Please also note that there are no camera poses for real-world objects from Pix3D and Stanford Cars. Our model assumes that objects are at the center of the canonical space and once trained, can estimate camera poses for each reference view similar to CodeNeRF~\cite{jang2021codenerf}.

\begin{table}[t]
\footnotesize
\centering 
	\caption{Quantitative comparisons against state-of-the-art methods on the $13$ largest categories of the ShapeNet-NMR dataset.}
	\resizebox{\linewidth}{!}{
		\renewcommand\arraystretch{0.85}	
		\begin{tabular}{cccccccccccccccc}
			\toprule
			& Methods & plane & bench & cbnt. & car & chair & disp. & lamp & spkr. & rifle & sofa & table & phone & boat & mean \\
			\midrule

			\multirow{4}{*}{PSNR$\uparrow$} & \cellcolor{mygray}DVR & \cellcolor{mygray}25.29 & \cellcolor{mygray}22.64 & \cellcolor{mygray}24.47 & \cellcolor{mygray}23.95 & \cellcolor{mygray}19.91 & \cellcolor{mygray}20.86 & \cellcolor{mygray}23.27 & \cellcolor{mygray}20.78 & \cellcolor{mygray}23.44 & \cellcolor{mygray}22.35 & \cellcolor{mygray}21.53 & \cellcolor{mygray}24.18 & \cellcolor{mygray}25.09 & \cellcolor{mygray}22.70 \\
			
			& SRN & 26.62 & 22.20 & 23.42 & 24.40 & 21.85 & 19.07 & 22.17 & 21.04 & 24.95 & 23.65 & 22.45 & 20.87 & 25.86 & 23.28 \\

			& \cellcolor{mygray}PixelNeRF & \cellcolor{mygray}29.76 & \cellcolor{mygray}26.35 & \cellcolor{mygray}27.72 & \cellcolor{mygray}27.58 & \cellcolor{mygray}23.84 & \cellcolor{mygray}24.22 & \cellcolor{mygray}28.58 & \cellcolor{mygray}24.44 & \cellcolor{mygray}30.60 & \cellcolor{mygray}26.94 & \cellcolor{mygray}25.59 & \cellcolor{mygray}27.13 & \cellcolor{mygray}29.18 & \cellcolor{mygray}26.80 \\
			
			& Ours & \textbf{30.57} & \textbf{27.44} & \textbf{29.34} & \textbf{27.87} & \textbf{24.29} & \textbf{24.90} & \textbf{28.98} & \textbf{25.14} & \textbf{30.64} & \textbf{27.70} & \textbf{27.16} & \textbf{28.27} & \textbf{29.71} & \textbf{27.57}\\
		
			\midrule
			\multirow{4}{*}{SSIM$\uparrow$} & \cellcolor{mygray}DVR & \cellcolor{mygray}0.905 & \cellcolor{mygray}0.866 & \cellcolor{mygray}0.877 & \cellcolor{mygray}0.909 & \cellcolor{mygray}0.787 & \cellcolor{mygray}0.814 & \cellcolor{mygray}0.849 & \cellcolor{mygray}0.798 & \cellcolor{mygray}0.916 & \cellcolor{mygray}0.868 & \cellcolor{mygray}0.840 & \cellcolor{mygray}0.892 & \cellcolor{mygray}0.902 & \cellcolor{mygray}0.860 \\
			
			& SRN & 0.901 & 0.837 & 0.831 & 0.897 & 0.814 & 0.744 & 0.801 & 0.779 & 0.913 & 0.851 & 0.828 & 0.811 & 0.898 & 0.849 \\
			
			& \cellcolor{mygray}PixelNeRF & \cellcolor{mygray}0.947 & \cellcolor{mygray}0.911 & \cellcolor{mygray}0.910 & \cellcolor{mygray}0.942 & \cellcolor{mygray}0.858 & \cellcolor{mygray}0.867 & \cellcolor{mygray}0.913 & \cellcolor{mygray}0.855 & \cellcolor{mygray}0.968 & \cellcolor{mygray}0.908 & \cellcolor{mygray}0.898 & \cellcolor{mygray}0.922 & \cellcolor{mygray}0.939 & \cellcolor{mygray}0.910 \\
			
			& Ours & \textbf{0.955} & \textbf{0.925} & \textbf{0.922} & \textbf{0.945} & \textbf{0.865} & \textbf{0.875} & \textbf{0.917} & \textbf{0.862} & \textbf{0.970} & \textbf{0.915} & \textbf{0.917} & \textbf{0.929} & \textbf{0.943} & \textbf{0.919} \\
			
			\midrule
			\multirow{4}{*}{LPIPS$\downarrow$} & \cellcolor{mygray}DVR & \cellcolor{mygray}0.095 & \cellcolor{mygray}0.129 & \cellcolor{mygray}0.125 & \cellcolor{mygray}0.098 & \cellcolor{mygray}0.173 & \cellcolor{mygray}0.150 & \cellcolor{mygray}0.172 & \cellcolor{mygray}0.170 & \cellcolor{mygray}0.094 & \cellcolor{mygray}0.119 & \cellcolor{mygray}0.139 & \cellcolor{mygray}0.110 & \cellcolor{mygray}0.116 & \cellcolor{mygray}0.130 \\
			
			& SRN & 0.111 & 0.150 & 0.147 & 0.115 & 0.152 & 0.197 & 0.210 & 0.178 & 0.111 & 0.129 & 0.135 & 0.165 & 0.134 & 0.139 \\
			
			& \cellcolor{mygray}PixelNeRF & \cellcolor{mygray}0.084 & \cellcolor{mygray}0.116 & \cellcolor{mygray}0.105 & \cellcolor{mygray}0.095 & \cellcolor{mygray}0.146 & \cellcolor{mygray}0.129 & \cellcolor{mygray}0.114 & \cellcolor{mygray}0.141 & \cellcolor{mygray}0.066 & \cellcolor{mygray}0.116 & \cellcolor{mygray}0.098 & \cellcolor{mygray}0.097 & \cellcolor{mygray}0.111 & \cellcolor{mygray}0.108 \\
			
			& Ours & \textbf{0.062} & \textbf{0.085} & \textbf{0.068} & \textbf{0.082} & \textbf{0.120} & \textbf{0.104} & \textbf{0.096} & \textbf{0.108} & \textbf{0.054} & \textbf{0.086} & \textbf{0.067} & \textbf{0.068} & \textbf{0.089} & \textbf{0.084} \\
			
			\bottomrule
		\end{tabular}
	}
	\label{tab:quantitative-nmr}
\end{table}

\begin{figure}[t]
    \centering
    \includegraphics[scale=0.15]{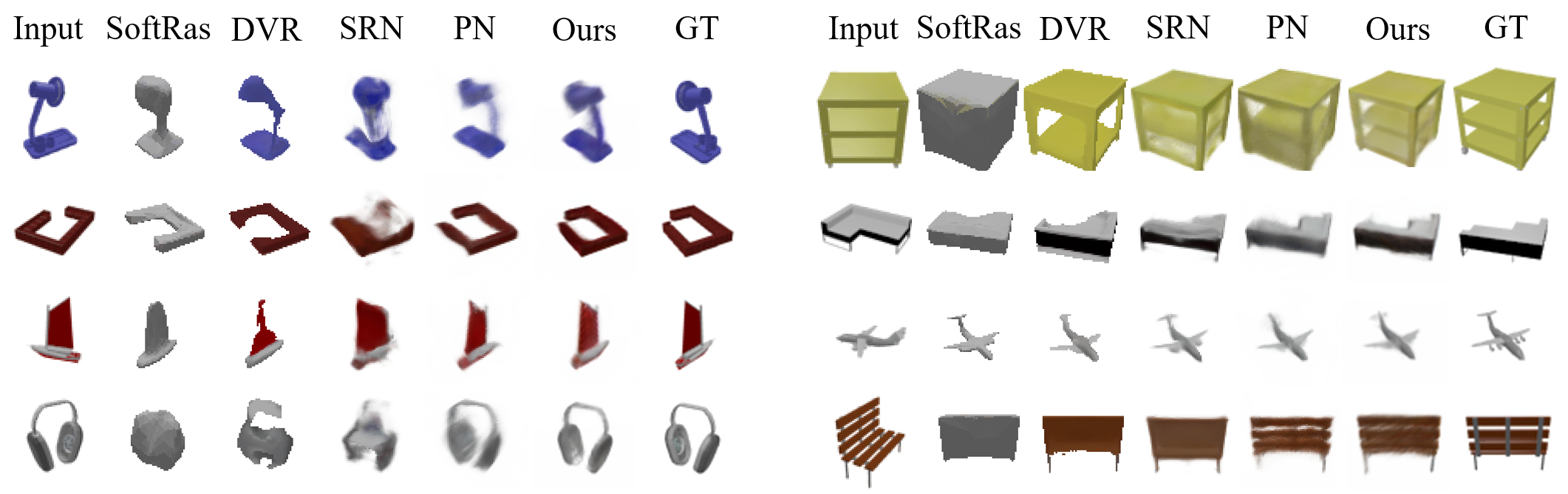}
    \caption{Qualitative comparisons on the ShapeNet-NMR dataset under the category-agnostic single-view reconstruction setting.}
    \label{fig:nmr}
\end{figure}

\begin{figure}[t]
    \centering
    \includegraphics[scale=0.18]{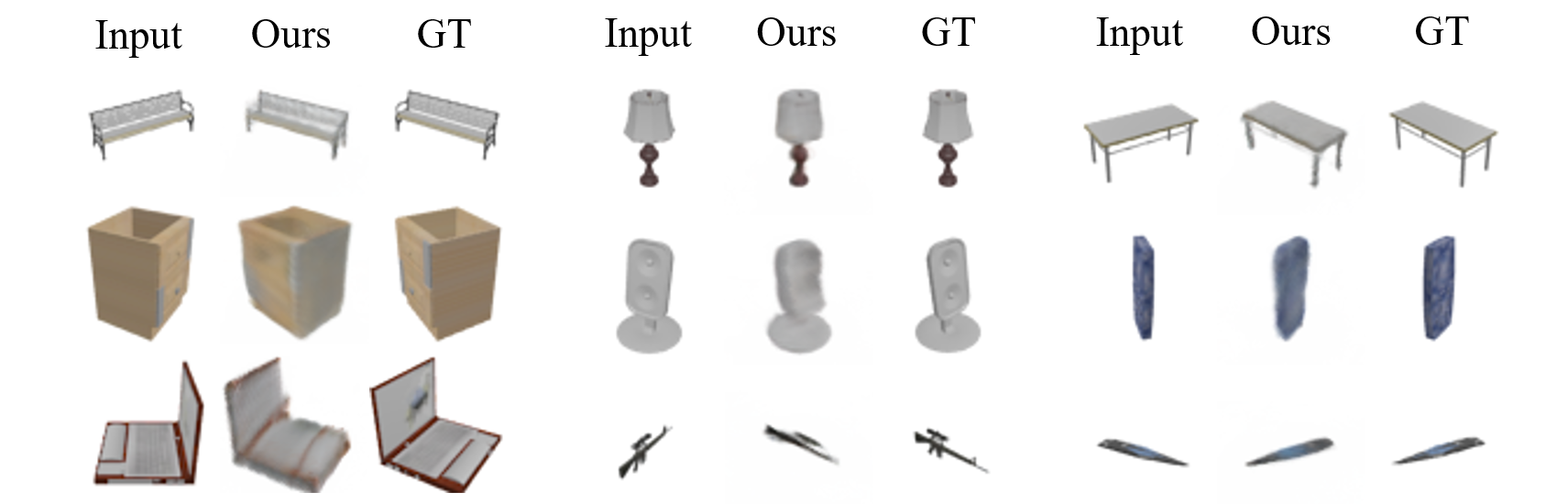}
    \caption{Qualitative visualization on the ShapeNet-NMR dataset under the generalization to unseen categories setting.}
    \label{fig:nmr-gen}
\end{figure}

\paragraph{\rm {\textbf{Evaluations on the ShapeNet-NMR Dataset.}}}
Although the experimental results of two common categories have demonstrated that including symmetry is simple yet effective, we further explore our approach on the ShapeNet-NMR dataset under two settings: category-agnostic single-view reconstruction and generalization to unseen categories. 1) Category-agnostic single-view reconstruction: only a single model is trained across the $13$ largest categories of ShapeNet. We show in Table~\ref{tab:quantitative-nmr} and Fig.~\ref{fig:nmr} that SymmNeRF outperforms other state-of-the-art methods~\cite{liu2019softras,niemeyer2020differentiable,sitzmann2019scene,yu2021pixelnerf}. This also implies that \textit{symmetry priors benefit the reconstruction of almost all symmetric objects}; 
2) Generalization to unseen categories: we reconstruct ShapeNet categories which are not involved in training. The results in Table~\ref{tab:quantitative-nmr-gen} and Fig.~\ref{fig:nmr-gen} suggest that our method performs comparably to PixelNeRF. This means that \textit{our method can also handle out-of-distribution categories with the help of symmetry priors.}

\begin{table}[t]
\footnotesize
\centering 
	\caption{Quantitative comparisons against state-of-the-art methods on $10$ unseen categories of ShapeNet-NMR dataset. The models are trained on only planes, cars and chairs.}
	\resizebox{0.9\linewidth}{!}{
		\renewcommand\arraystretch{0.85}	
		\begin{tabular}{ccccccccccccc}
			\toprule
			& Methods & bench & cbnt. & disp. & lamp & spkr. & rifle & sofa & table & phone & boat & mean \\
			\midrule

			\multirow{4}{*}{PSNR$\uparrow$} & \cellcolor{mygray}DVR & \cellcolor{mygray}18.37 & \cellcolor{mygray}17.19 & \cellcolor{mygray}14.33 & \cellcolor{mygray}18.48 & \cellcolor{mygray}16.09 & \cellcolor{mygray}20.28 & \cellcolor{mygray}18.62 & \cellcolor{mygray}16.20 & \cellcolor{mygray}16.84 & \cellcolor{mygray}22.43 & \cellcolor{mygray}17.72 \\
			
			& SRN & 18.71 & 17.04 & 15.06 & 19.26 & 17.06 & 23.12 & 18.76 & 17.35 & 15.66 & 24.97 & 18.71 \\

			& \cellcolor{mygray}PixelNeRF & \cellcolor{mygray}23.79 & \cellcolor{mygray}\textbf{22.85} & \cellcolor{mygray}\textbf{18.09} & \cellcolor{mygray}\textbf{22.76} & \cellcolor{mygray}\textbf{21.22} & \cellcolor{mygray}23.68 & \cellcolor{mygray}24.62 & \cellcolor{mygray}\textbf{21.65} & \cellcolor{mygray}\textbf{21.05} & \cellcolor{mygray}26.55 & \cellcolor{mygray}\textbf{22.71} \\
			
			& Ours & \textbf{23.87} & 21.36 & 16.83 & 22.68 & 19.98 & \textbf{23.77} & \textbf{25.10} & 21.10 & 20.48 & \textbf{26.80} & 22.36 \\
		
			\midrule
			\multirow{4}{*}{SSIM$\uparrow$} & \cellcolor{mygray}DVR & \cellcolor{mygray}0.754 & \cellcolor{mygray}0.686 & \cellcolor{mygray}0.601 & \cellcolor{mygray}0.749 & \cellcolor{mygray}0.657 & \cellcolor{mygray}0.858 & \cellcolor{mygray}0.755 & \cellcolor{mygray}0.644 & \cellcolor{mygray}0.731 & \cellcolor{mygray}0.857 & \cellcolor{mygray}0.716 \\
			
			& SRN & 0.702 & 0.626 & 0.577 & 0.685 & 0.633 & 0.875 & 0.702 & 0.617 & 0.635 & 0.875 & 0.684 \\
			
			& \cellcolor{mygray}PixelNeRF & \cellcolor{mygray}0.863 & \cellcolor{mygray}\textbf{0.814} & \cellcolor{mygray}\textbf{0.687} & \cellcolor{mygray}0.818 & \cellcolor{mygray}\textbf{0.778} & \cellcolor{mygray}0.899 & \cellcolor{mygray}0.866 & \cellcolor{mygray}\textbf{0.798} & \cellcolor{mygray}0.801 & \cellcolor{mygray}0.896 & \cellcolor{mygray}\textbf{0.825} \\
			
			& Ours & \textbf{0.873} & 0.780 & 0.663 & \textbf{0.824} & 0.751 & \textbf{0.902} & \textbf{0.881} & 0.792 & \textbf{0.802} & \textbf{0.909} & 0.823 \\
			
			\midrule
			\multirow{4}{*}{LPIPS$\downarrow$} & \cellcolor{mygray}DVR & \cellcolor{mygray}0.219 & \cellcolor{mygray}0.257 & \cellcolor{mygray}0.306 & \cellcolor{mygray}0.259 & \cellcolor{mygray}0.266 & \cellcolor{mygray}0.158 & \cellcolor{mygray}0.196 & \cellcolor{mygray}0.280 & \cellcolor{mygray}0.245 & \cellcolor{mygray}0.152 & \cellcolor{mygray}0.240 \\
			
			& SRN & 0.282 & 0.314 & 0.333 & 0.321 & 0.289 & 0.175 & 0.248 & 0.315 & 0.324 & 0.163 & 0.280 \\
			
			& \cellcolor{mygray}PixelNeRF & \cellcolor{mygray}0.164 & \cellcolor{mygray}0.186 & \cellcolor{mygray}0.271 & \cellcolor{mygray}0.208 & \cellcolor{mygray}0.203 & \cellcolor{mygray}0.141 & \cellcolor{mygray}0.157 & \cellcolor{mygray}0.188 & \cellcolor{mygray}0.207 & \cellcolor{mygray}0.148 & \cellcolor{mygray}0.182 \\
			
			& Ours & \textbf{0.126} & \textbf{0.174} & \textbf{0.251} & \textbf{0.184} & \textbf{0.185} & \textbf{0.121} & \textbf{0.115} & \textbf{0.163} & \textbf{0.178} & \textbf{0.111} & \textbf{0.155} \\
			
			\bottomrule
		\end{tabular}
	}
	\label{tab:quantitative-nmr-gen}
\end{table}

\begin{table}[t]
\centering
    \caption{Ablation study on each component of SymmNeRF.}
	\resizebox{0.75\linewidth}{!}{
		\renewcommand\arraystretch{1.00}	
		\begin{tabular}{cccccccc}
			\toprule
			& & \makecell[c]{Image encoder \\ network} & \makecell[c]{Hypernetwork} & \makecell[c]{Local \\ features} & \makecell[c]{Symm \\ features} & PSNR($\Delta$)$\uparrow$ & SSIM($\Delta$)$\uparrow$ \\
			\midrule

			\multirow{4}{*}{Chairs} & (a) & $\checkmark$ & $\checkmark$ & & & 21.26 (-) & 0.87 (-) \\
			
			& (b) & $\checkmark$ & $\checkmark$ & $\checkmark$ & & 23.09 (8.6\%) & 0.91 (4.6\%) \\
			
			& (c) & $\checkmark$ & $\checkmark$ & $\checkmark$ & $\checkmark$ & \textbf{24.32 (14.4\%)} & \textbf{0.92 (5.7\%)} \\
			
			& (d) & $\checkmark$ & & $\checkmark$ & $\checkmark$ & 19.76 (-7.1\%) & 0.85 (-2.3\%) \\

			\midrule
			\multirow{4}{*}{Cars} & (a) & $\checkmark$ & $\checkmark$ & & & 20.65 (-) & 0.87 (-) \\
			
			& (b) & $\checkmark$ & $\checkmark$ & $\checkmark$ & & 22.15 (7.3\%) & 0.89 (2.3\%) \\
			
			& (c) & $\checkmark$ & $\checkmark$ & $\checkmark$ & $\checkmark$ & \textbf{23.44 (13.5\%)} & \textbf{0.91 (4.6\%)} \\
			
			& (d) & $\checkmark$ & & $\checkmark$ & $\checkmark$ & 21.15 (2.4\%) & 0.86 (-1.2\%) \\
			
			\midrule
			\multirow{4}{*}{Average} & (a) & $\checkmark$ & $\checkmark$ & & & 20.96 (-) & 0.87 (-) \\
			
			& (b) & $\checkmark$ & $\checkmark$ & $\checkmark$ & & 22.62 (7.9\%) & 0.90 (3.4\%) \\
			
			& (c) & $\checkmark$ & $\checkmark$ & $\checkmark$ & $\checkmark$ & \textbf{23.88 (13.9\%)} & \textbf{0.92 (5.7\%)} \\
			
			& (d) & $\checkmark$ & & $\checkmark$ & $\checkmark$ & 20.46 (-2.4\%) & 0.86 (-1.1\%) \\
			\bottomrule
			\vspace{-5mm}
		\end{tabular}
	}
	\label{tab:ablation}
\end{table}

\paragraph{\rm {\textbf{Asymmetric Objects.}}} 
As shown in Fig.~\ref{fig:nmr} (Row 2), \textit{our method can also deal with objects that are not perfectly symmetric.} This is because a few asymmetric objects are also included in the training dataset. Our model can perceive and recognize asymmetry thanks to the global latent code and hypernetwork. SymmNeRF therefore adaptively chooses to utilize local features to reconstruct asymmetric objects.


\subsection{Ablation Study}
\begin{figure}[t]
    \centering
    \includegraphics[scale=0.14]{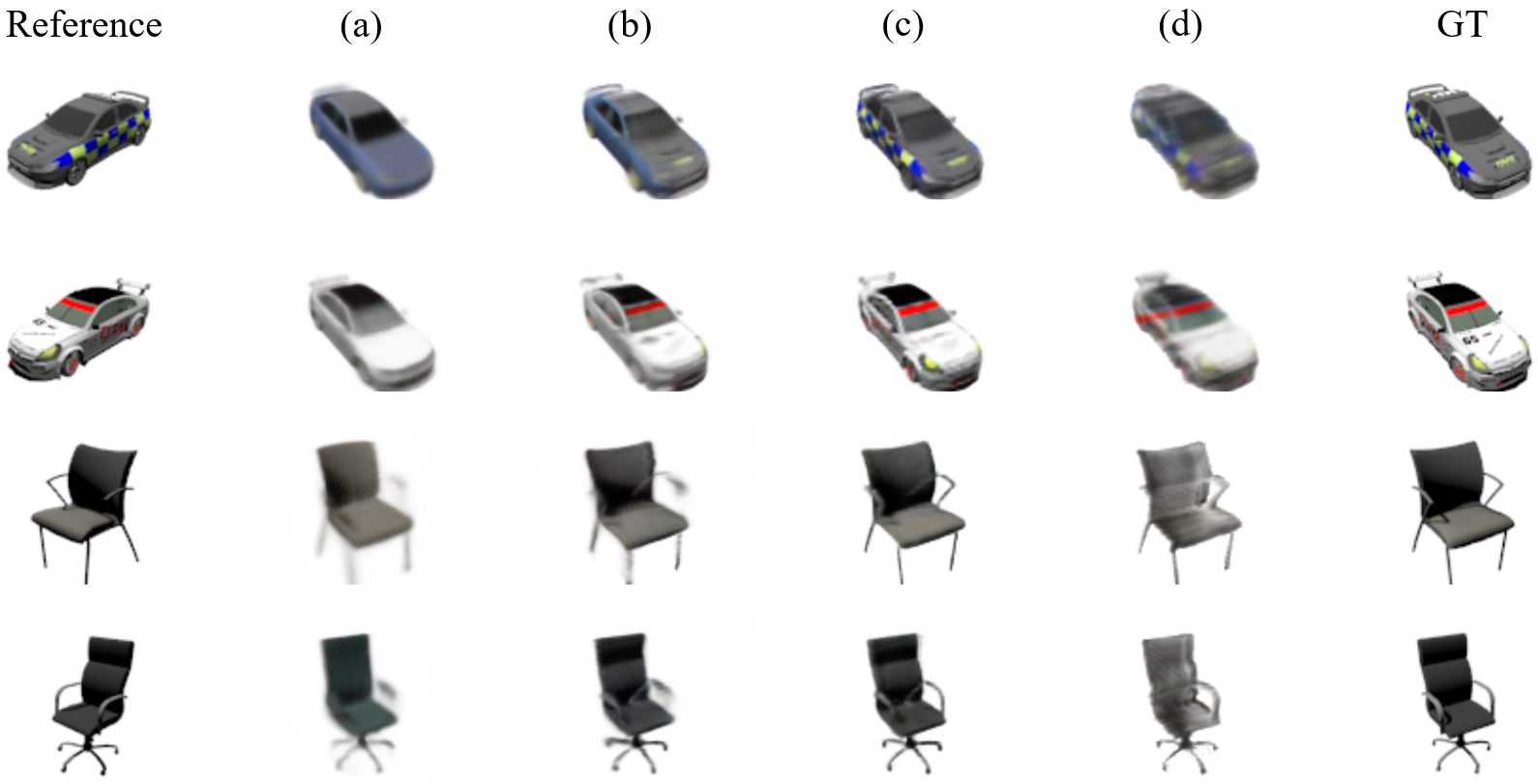}
    \caption{Qualitative evaluation of different configurations on ShapeNet-SRN.}
    \vspace{-5mm}
    \label{fig:ablation}
\end{figure}

To validate the design choice of SymmNeRF, we conduct ablation studies on the synthetic ``Chairs'' and ``Cars'' from the ShapeNet-SRN dataset. 
Table~\ref{tab:ablation} shows the results corresponding to the effectiveness of the pixel-aligned, symmetric features and the hypernetwork. 
One can observe: 
i) \textit{The symmetry priors injection benefits novel view synthesis.} Compared with (b) in average performance, our full model (c) with the symmetric priors injection yields a relative improvement of $6.0$\% PSNR and $2.3$\% SSIM. This finding highlights the importance of the symmetry priors on novel view synthesis when only a single image is provided; ii) \textit{The hypernetwork matters.} Compared with our full model (c), the rendering quality of (d) deteriorates if we do not adopt the hypernetwork. This may lie in the fact that simply conditioning on local features ignores the underlying 3D structure of objects. In contrast, combining local and global conditioning via the hypernetwork module not only enables recovery of rendering details, but also improves generalization to unseen objects in a coarse-to-fine manner.
We also visualize the comparative results in Fig.~\ref{fig:ablation}. 
The baseline model (a) tends to render smoothly. Simply using pixel-aligned image features (b) still fails to fully understand 3D structure. In contrast, our full model (c) reproduces photo-realistic details from most viewpoints. The rendering quality of (d) deteriorates as the hypernetwork is not adopted. 
We have to emphasize that, \textit{only including both the symmetry priors and the hypernetwork can accurately recovers the geometry information and texture details despite the occlusions.}

\section{Conclusion}
Existing methods~\cite{jang2021codenerf,yu2021pixelnerf} fail to synthesize fine appearance details of objects, especially when the target view is far away from the reference view. They focus on learning scene priors, but ignore fully exploring the attributes of objects, \textit{e.g.}, symmetry. In this paper, we investigate the potential performance gains of explicitly injecting symmetry priors into the scene representation. 
In particular, we combine hypernetworks~\cite{sitzmann2019scene} with local conditioning~\cite{saito2019pifu,Trevithick_2021_ICCV,yu2021pixelnerf}, embedded with the symmetry prior. Experimental results demonstrate that such a symmetry prior can boost our model to synthesize novel views with more details regardless of the pose transformation, and show good generalization when applied to unseen objects. 

\subsubsection{Acknowledgement.}
This work is supported in part by the National Natural Science Foundation of China (Grant No. U1913602). This study is also supported under the RIE2020 Industry Alignment Fund – Industry Collaboration Projects (IAF-ICP) Funding Initiative, as well as cash and in-kind contribution from the industry partner(s).

%
%
\bibliographystyle{splncs04}
\bibliography{main}

\appendix   

\setcounter{table}{0}   
\setcounter{figure}{0}

\renewcommand\thesection{A\arabic{section}}
\renewcommand\thesubsection{A\arabic{subsection}}
\renewcommand\thefigure{A\arabic{figure}}
\renewcommand\theequation{A\arabic{equation}}
\renewcommand\thetable{A\arabic{table}}

\section*{Appendix}

\subsection{Qualitative Comparisons on the ShapeNet-SRN Dataset}
\begin{figure*}
    \centering
    \begin{minipage}{\textwidth}
        \begin{minipage}{0.12\textwidth}
            \centering
            \subcaption*{Reference}
            \includegraphics[scale=0.35]{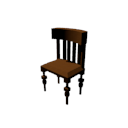} \\
            \includegraphics[scale=0.35]{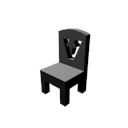} \\
            \includegraphics[scale=0.35]{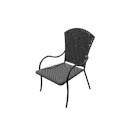} \\
            \includegraphics[scale=0.35]{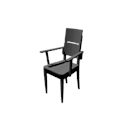} \\
            \includegraphics[scale=0.35]{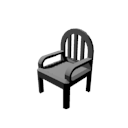} \\
            \includegraphics[scale=0.35]{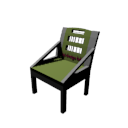} \\
            \includegraphics[scale=0.35]{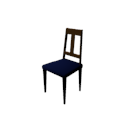} \\
            \includegraphics[scale=0.35]{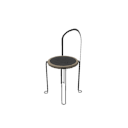} \\
            \includegraphics[scale=0.35]{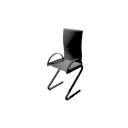} \\
            \includegraphics[scale=0.35]{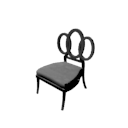}
        \end{minipage}
        \begin{minipage}{0.88\textwidth}
            \begin{minipage}{0.25\textwidth}
                \centering
                \subcaption*{SRN}
                \begin{minipage}{0.5\textwidth}
                    \includegraphics[scale=0.35]{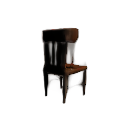} \\
                    \includegraphics[scale=0.35]{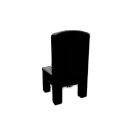} \\
                    \includegraphics[scale=0.35]{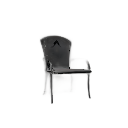} \\
                    \includegraphics[scale=0.35]{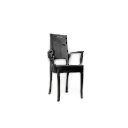} \\
                    \includegraphics[scale=0.35]{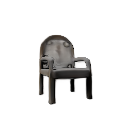} \\
                    \includegraphics[scale=0.35]{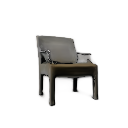} \\
                    \includegraphics[scale=0.35]{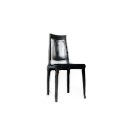} \\
                    \includegraphics[scale=0.35]{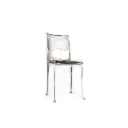} \\
                    \includegraphics[scale=0.35]{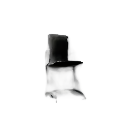} \\
                    \includegraphics[scale=0.35]{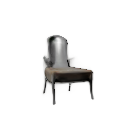}
                \end{minipage}
                \hspace{-3mm}
                \begin{minipage}{0.5\textwidth}
                    \includegraphics[scale=0.35]{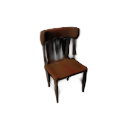} \\
                    \includegraphics[scale=0.35]{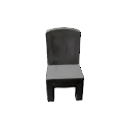} \\
                    \includegraphics[scale=0.35]{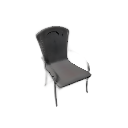} \\
                    \includegraphics[scale=0.35]{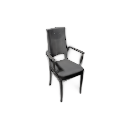} \\
                    \includegraphics[scale=0.35]{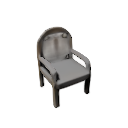} \\
                    \includegraphics[scale=0.35]{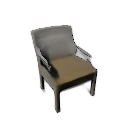} \\
                    \includegraphics[scale=0.35]{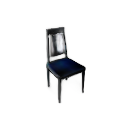} \\
                    \includegraphics[scale=0.35]{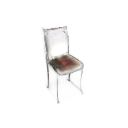} \\
                    \includegraphics[scale=0.35]{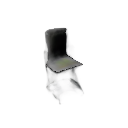} \\
                    \includegraphics[scale=0.35]{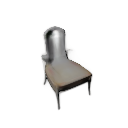}
                \end{minipage}
            \end{minipage}
            \hspace{-3mm}
            \begin{minipage}{0.25\textwidth}
                \centering
                \subcaption*{PixelNeRF}
                \begin{minipage}{0.5\textwidth}
                    \includegraphics[scale=0.35]{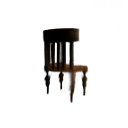} \\
                    \includegraphics[scale=0.35]{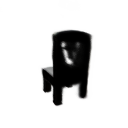} \\
                    \includegraphics[scale=0.35]{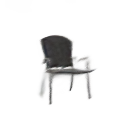} \\
                    \includegraphics[scale=0.35]{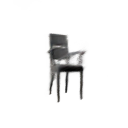} \\
                    \includegraphics[scale=0.35]{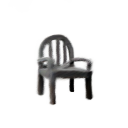} \\
                    \includegraphics[scale=0.35]{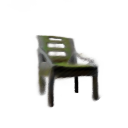} \\
                    \includegraphics[scale=0.35]{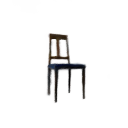} \\
                    \includegraphics[scale=0.35]{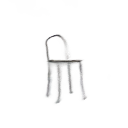} \\
                    \includegraphics[scale=0.35]{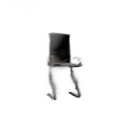} \\
                    \includegraphics[scale=0.35]{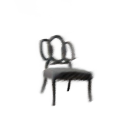}
                \end{minipage}
                \hspace{-3mm}
                \begin{minipage}{0.5\textwidth}
                    \includegraphics[scale=0.35]{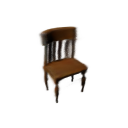} \\
                    \includegraphics[scale=0.35]{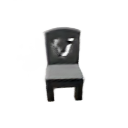} \\
                    \includegraphics[scale=0.35]{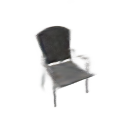} \\
                    \includegraphics[scale=0.35]{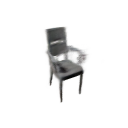} \\
                    \includegraphics[scale=0.35]{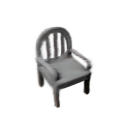} \\
                    \includegraphics[scale=0.35]{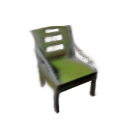} \\
                    \includegraphics[scale=0.35]{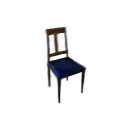} \\
                    \includegraphics[scale=0.35]{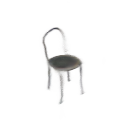} \\
                    \includegraphics[scale=0.35]{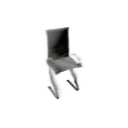} \\
                    \includegraphics[scale=0.35]{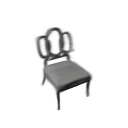}
                \end{minipage}
            \end{minipage}
            \hspace{-3mm}
            \begin{minipage}{0.25\textwidth}
                \centering
                \subcaption*{Ours}
                \begin{minipage}{0.5\textwidth}
                    \includegraphics[scale=0.35]{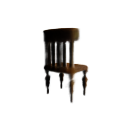} \\
                    \includegraphics[scale=0.35]{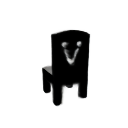} \\
                    \includegraphics[scale=0.35]{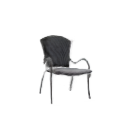} \\
                    \includegraphics[scale=0.35]{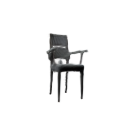} \\
                    \includegraphics[scale=0.35]{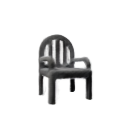} \\
                    \includegraphics[scale=0.35]{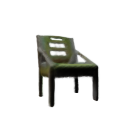} \\
                    \includegraphics[scale=0.35]{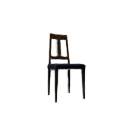} \\
                    \includegraphics[scale=0.35]{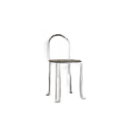} \\
                    \includegraphics[scale=0.35]{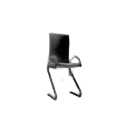} \\
                    \includegraphics[scale=0.35]{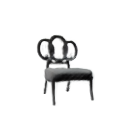}
                \end{minipage}
                \hspace{-3mm}
                \begin{minipage}{0.5\textwidth}
                    \includegraphics[scale=0.35]{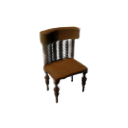} \\
                    \includegraphics[scale=0.35]{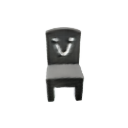} \\
                    \includegraphics[scale=0.35]{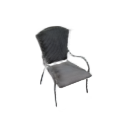} \\
                    \includegraphics[scale=0.35]{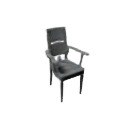} \\
                    \includegraphics[scale=0.35]{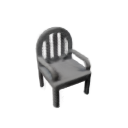} \\
                    \includegraphics[scale=0.35]{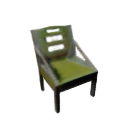} \\
                    \includegraphics[scale=0.35]{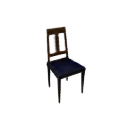} \\
                    \includegraphics[scale=0.35]{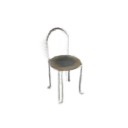} \\
                    \includegraphics[scale=0.35]{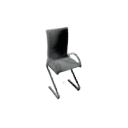} \\
                    \includegraphics[scale=0.35]{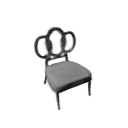}
                \end{minipage}
            \end{minipage}
            \hspace{-3mm}
            \begin{minipage}{0.25\textwidth}
                \centering
                \subcaption*{GT}
                \begin{minipage}{0.5\textwidth}
                    \includegraphics[scale=0.35]{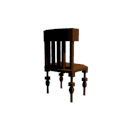} \\
                    \includegraphics[scale=0.35]{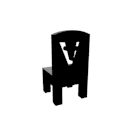} \\
                    \includegraphics[scale=0.35]{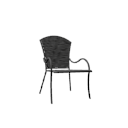} \\
                    \includegraphics[scale=0.35]{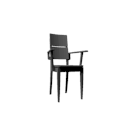} \\
                    \includegraphics[scale=0.35]{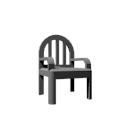} \\
                    \includegraphics[scale=0.35]{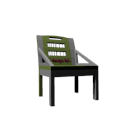} \\
                    \includegraphics[scale=0.35]{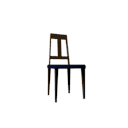} \\
                    \includegraphics[scale=0.35]{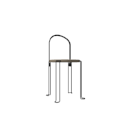} \\
                    \includegraphics[scale=0.35]{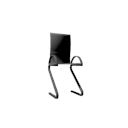} \\
                    \includegraphics[scale=0.35]{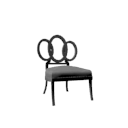}
                \end{minipage}
                \hspace{-3mm}
                \begin{minipage}{0.5\textwidth}
                    \includegraphics[scale=0.35]{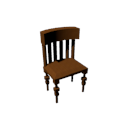} \\
                    \includegraphics[scale=0.35]{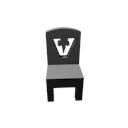} \\
                    \includegraphics[scale=0.35]{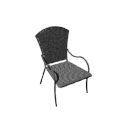} \\
                    \includegraphics[scale=0.35]{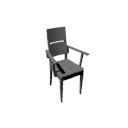} \\
                    \includegraphics[scale=0.35]{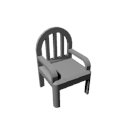} \\
                    \includegraphics[scale=0.35]{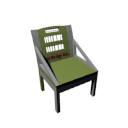} \\
                    \includegraphics[scale=0.35]{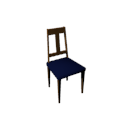} \\
                    \includegraphics[scale=0.35]{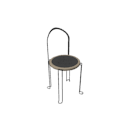} \\
                    \includegraphics[scale=0.35]{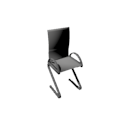} \\
                    \includegraphics[scale=0.35]{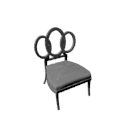}
                \end{minipage}
            \end{minipage}
        \end{minipage}
    \end{minipage}
    \caption{Additional qualitative comparisons on the ``Chairs'' category of the ShapeNet-SRN~\cite{chang2015shapenet,sitzmann2019scene} dataset. Compared with SRN~\cite{sitzmann2019scene} and PixelNeRF~\cite{yu2021pixelnerf}, SymmNeRF yields more photo-realistic and reasonable novel views with fine-grained details close to ground truths.} 
    \label{fig:qualititive-chairs}
\end{figure*}

\begin{figure*}
    \centering
    \begin{minipage}{\textwidth}
        \begin{minipage}{0.12\textwidth}
            \centering
            \subcaption*{Reference}
            \includegraphics[scale=0.3]{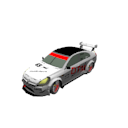} \\
            \includegraphics[scale=0.3]{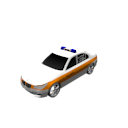} \\
            \includegraphics[scale=0.3]{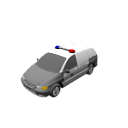} \\
            \includegraphics[scale=0.3]{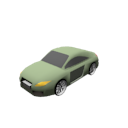} \\
            \includegraphics[scale=0.3]{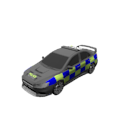} \\
            \includegraphics[scale=0.3]{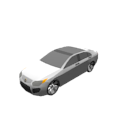} \\
            \includegraphics[scale=0.3]{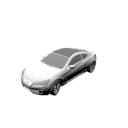} \\
            \includegraphics[scale=0.3]{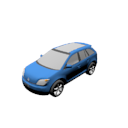} \\
            \includegraphics[scale=0.3]{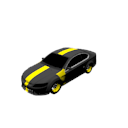} \\
            \includegraphics[scale=0.3]{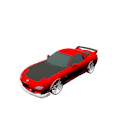}
        \end{minipage}
        \begin{minipage}{0.88\textwidth}
            \begin{minipage}{0.25\textwidth}
                \centering
                \vspace{-1.2mm}
                \subcaption*{SRN}
                \vspace{0.8mm}
                \begin{minipage}{0.5\textwidth}
                    \centering
                    \includegraphics[scale=0.25]{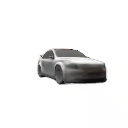} \\
                    \vspace{2.3mm}
                    \includegraphics[scale=0.25]{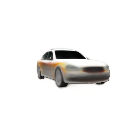} \\
                    \vspace{2.3mm}
                    \includegraphics[scale=0.25]{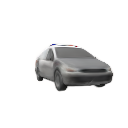} \\
                    \vspace{2.3mm}
                    \includegraphics[scale=0.25]{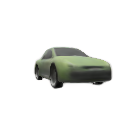} \\
                    \vspace{2.3mm}
                    \includegraphics[scale=0.25]{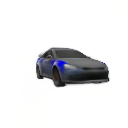} \\
                    \vspace{2.3mm}
                    \includegraphics[scale=0.25]{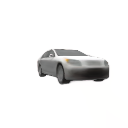} \\
                    \vspace{2.3mm}
                    \includegraphics[scale=0.25]{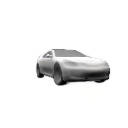} \\
                    \vspace{2.3mm}
                    \includegraphics[scale=0.25]{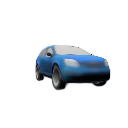} \\
                    \vspace{2.3mm}
                    \includegraphics[scale=0.25]{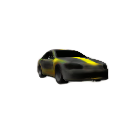} \\
                    \vspace{2.3mm}
                    \includegraphics[scale=0.25]{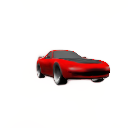}
                \end{minipage}
                \hspace{-3mm}
                \begin{minipage}{0.5\textwidth}
                    \centering
                    \includegraphics[scale=0.25]{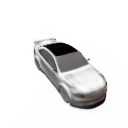} \\
                    \vspace{2.3mm}
                    \includegraphics[scale=0.25]{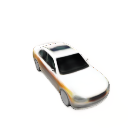} \\
                    \vspace{2.3mm}
                    \includegraphics[scale=0.25]{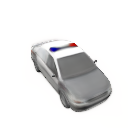} \\
                    \vspace{2.3mm}
                    \includegraphics[scale=0.25]{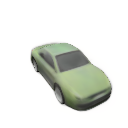} \\
                    \vspace{2.3mm}
                    \includegraphics[scale=0.25]{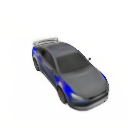} \\
                    \vspace{2.3mm}
                    \includegraphics[scale=0.25]{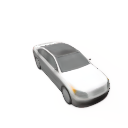} \\
                    \vspace{2.3mm}
                    \includegraphics[scale=0.25]{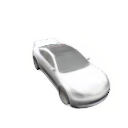} \\
                    \vspace{2.3mm}
                    \includegraphics[scale=0.25]{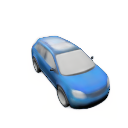} \\
                    \vspace{2.3mm}
                    \includegraphics[scale=0.25]{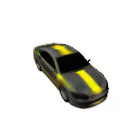} \\
                    \vspace{2.3mm}
                    \includegraphics[scale=0.25]{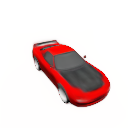}
                \end{minipage}
            \end{minipage}
            \hspace{-3mm}
            \begin{minipage}{0.25\textwidth}
                \centering
                \subcaption*{PixelNeRF}
                \begin{minipage}{0.5\textwidth}
                    \centering
                    \includegraphics[scale=0.3]{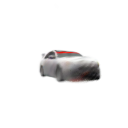} \\
                    \includegraphics[scale=0.3]{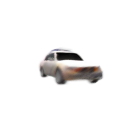} \\
                    \includegraphics[scale=0.3]{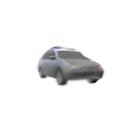} \\
                    \includegraphics[scale=0.3]{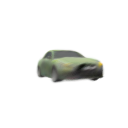} \\
                    \includegraphics[scale=0.3]{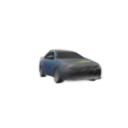} \\
                    \includegraphics[scale=0.3]{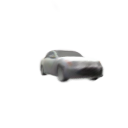} \\
                    \includegraphics[scale=0.3]{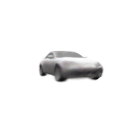} \\
                    \includegraphics[scale=0.3]{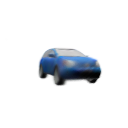} \\
                    \includegraphics[scale=0.3]{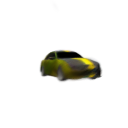} \\
                    \includegraphics[scale=0.3]{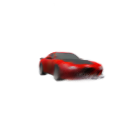}
                \end{minipage}
                \hspace{-3mm}
                \begin{minipage}{0.5\textwidth}
                    \centering
                    \includegraphics[scale=0.3]{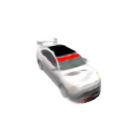} \\
                    \includegraphics[scale=0.3]{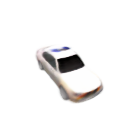} \\
                    \includegraphics[scale=0.3]{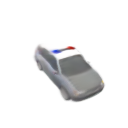} \\
                    \includegraphics[scale=0.3]{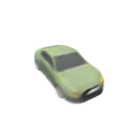} \\
                    \includegraphics[scale=0.3]{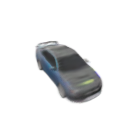} \\
                    \includegraphics[scale=0.3]{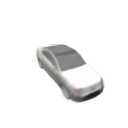} \\
                    \includegraphics[scale=0.3]{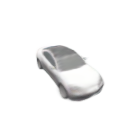} \\
                    \includegraphics[scale=0.3]{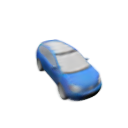} \\
                    \includegraphics[scale=0.3]{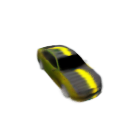} \\
                    \includegraphics[scale=0.3]{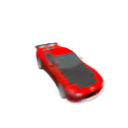}
                \end{minipage}
            \end{minipage}
            \hspace{-3mm}
            \begin{minipage}{0.25\textwidth}
                \centering
                \subcaption*{Ours}
                \begin{minipage}{0.5\textwidth}
                    \includegraphics[scale=0.3]{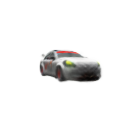} \\
                    \includegraphics[scale=0.3]{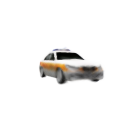} \\
                    \includegraphics[scale=0.3]{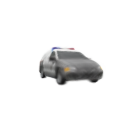} \\
                    \includegraphics[scale=0.3]{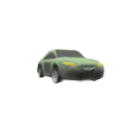} \\
                    \includegraphics[scale=0.3]{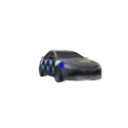} \\
                    \includegraphics[scale=0.3]{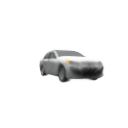} \\
                    \includegraphics[scale=0.3]{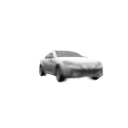} \\
                    \includegraphics[scale=0.3]{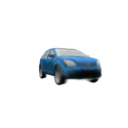} \\
                    \includegraphics[scale=0.3]{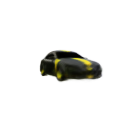} \\
                    \includegraphics[scale=0.3]{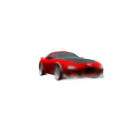}
                \end{minipage}
                \hspace{-3mm}
                \begin{minipage}{0.5\textwidth}
                    \includegraphics[scale=0.3]{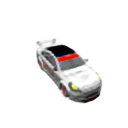} \\
                    \includegraphics[scale=0.3]{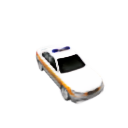} \\
                    \includegraphics[scale=0.3]{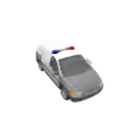} \\
                    \includegraphics[scale=0.3]{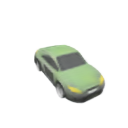} \\
                    \includegraphics[scale=0.3]{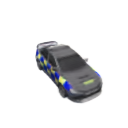} \\
                    \includegraphics[scale=0.3]{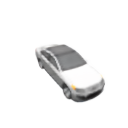} \\
                    \includegraphics[scale=0.3]{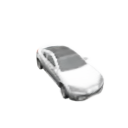} \\
                    \includegraphics[scale=0.3]{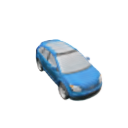} \\
                    \includegraphics[scale=0.3]{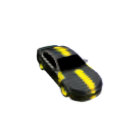} \\
                    \includegraphics[scale=0.3]{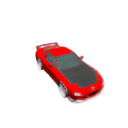}
                \end{minipage}
            \end{minipage}
            \hspace{-3mm}
            \begin{minipage}{0.25\textwidth}
                \centering
                \subcaption*{GT}
                \begin{minipage}{0.5\textwidth}
                    \includegraphics[scale=0.3]{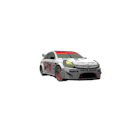} \\
                    \includegraphics[scale=0.3]{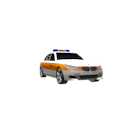} \\
                    \includegraphics[scale=0.3]{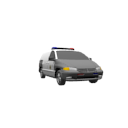} \\
                    \includegraphics[scale=0.3]{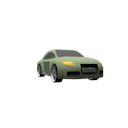} \\
                    \includegraphics[scale=0.3]{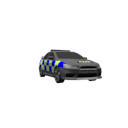} \\
                    \includegraphics[scale=0.3]{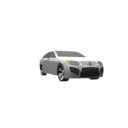} \\
                    \includegraphics[scale=0.3]{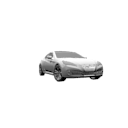} \\
                    \includegraphics[scale=0.3]{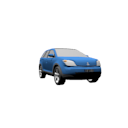} \\
                    \includegraphics[scale=0.3]{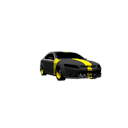} \\
                    \includegraphics[scale=0.3]{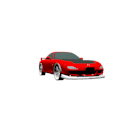}
                \end{minipage}
                \hspace{-3mm}
                \begin{minipage}{0.5\textwidth}
                    \includegraphics[scale=0.3]{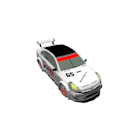} \\
                    \includegraphics[scale=0.3]{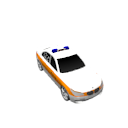} \\
                    \includegraphics[scale=0.3]{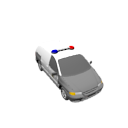} \\
                    \includegraphics[scale=0.3]{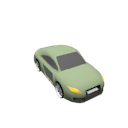} \\
                    \includegraphics[scale=0.3]{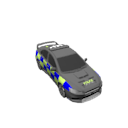} \\
                    \includegraphics[scale=0.3]{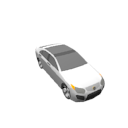} \\
                    \includegraphics[scale=0.3]{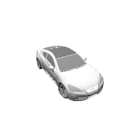} \\
                    \includegraphics[scale=0.3]{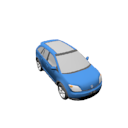} \\
                    \includegraphics[scale=0.3]{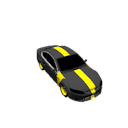} \\
                    \includegraphics[scale=0.3]{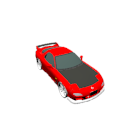}
                \end{minipage}
            \end{minipage}
        \end{minipage}
    \end{minipage}
    \caption{Additional qualitative comparisons on the ``Cars'' category of the ShapeNet-SRN~\cite{chang2015shapenet,sitzmann2019scene} dataset. Compared with SRN~\cite{sitzmann2019scene} and PixelNeRF~\cite{yu2021pixelnerf}, SymmNeRF yields more photo-realistic and reasonable novel views with fine-grained details close to ground truths.} 
    \label{fig:qualititive-cars}
\end{figure*}

As shown in Fig.~\ref{fig:qualititive-chairs} and Fig.~\ref{fig:qualititive-cars}, compared with SRN~\cite{sitzmann2019scene} and PixelNeRF~\cite{yu2021pixelnerf}, our method can synthesize more photo-realistic and reasonable novel views with fine-grained details close to ground truths.

\subsection{Novel View Synthesis on the ShapeNet-SRN Dataset}
\begin{figure*}
    \centering
    \begin{minipage}{\textwidth}
        \begin{minipage}{0.12\textwidth}
            \centering
            \subcaption*{Reference}
            \includegraphics[scale=0.3]{figures/qualitative/chairs/e53b07b648e8d041107a17cfae0b6df6/src_rgb.png} \\
            \includegraphics[scale=0.3]{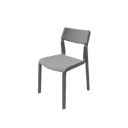} \\        \includegraphics[scale=0.3]{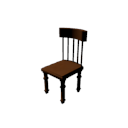} \\
            \includegraphics[scale=0.3]{figures/qualitative/chairs/ef3377832d90dbbacfe150564cb24aad/src_rgb.png} \\
            \includegraphics[scale=0.3]{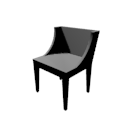} \\
            \includegraphics[scale=0.3]{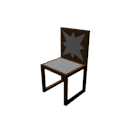} \\
            \includegraphics[scale=0.3]{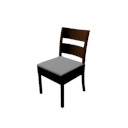} \\
            \includegraphics[scale=0.3]{figures/qualitative/chairs/9d395454d6de675d2025ebfdd95f4ba7/src_rgb.png} \\
            \includegraphics[scale=0.3]{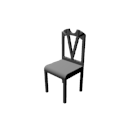} \\
            \includegraphics[scale=0.3]{figures/qualitative/chairs/c7e590c0390e8d5debe67d9b32c3ddf8/src_rgb.png}
        \end{minipage}
        \hspace{1mm}
        \vline
        \hspace{-3mm}
        \begin{minipage}{0.88\textwidth}
            \centering
            \subcaption*{Novel Views}
            \begin{minipage}{0.1\textwidth}
                \includegraphics[scale=0.3]{figures/qualitative/chairs/e53b07b648e8d041107a17cfae0b6df6/000000_pred_coarse.png} \\
                \includegraphics[scale=0.3]{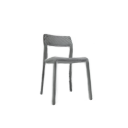} \\
                \includegraphics[scale=0.3]{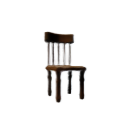} \\
                \includegraphics[scale=0.3]{figures/qualitative/chairs/ef3377832d90dbbacfe150564cb24aad/000000_pred_coarse.png} \\
                \includegraphics[scale=0.3]{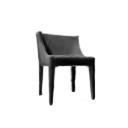} \\
                \includegraphics[scale=0.3]{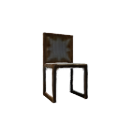} \\
                \includegraphics[scale=0.3]{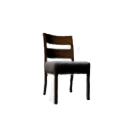} \\
                \includegraphics[scale=0.3]{figures/qualitative/chairs/9d395454d6de675d2025ebfdd95f4ba7/000000_pred_coarse.png} \\
                \includegraphics[scale=0.3]{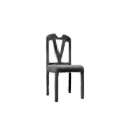} \\
                \includegraphics[scale=0.3]{figures/qualitative/chairs/c7e590c0390e8d5debe67d9b32c3ddf8/000000_pred_coarse.png}
            \end{minipage}
            \hspace{-1.5mm}
            \begin{minipage}{0.1\textwidth}
                \includegraphics[scale=0.3]{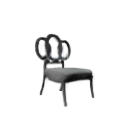} \\
                \includegraphics[scale=0.3]{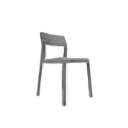} \\
                \includegraphics[scale=0.3]{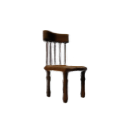} \\
                \includegraphics[scale=0.3]{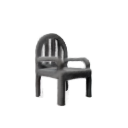} \\
                \includegraphics[scale=0.3]{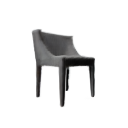} \\
                \includegraphics[scale=0.3]{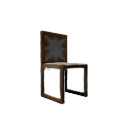} \\
                \includegraphics[scale=0.3]{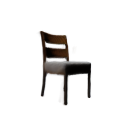} \\
                \includegraphics[scale=0.3]{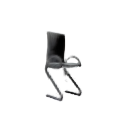} \\
                \includegraphics[scale=0.3]{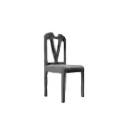} \\
                \includegraphics[scale=0.3]{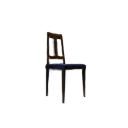}
            \end{minipage}
            \hspace{-1.5mm}
            \begin{minipage}{0.1\textwidth}
                \includegraphics[scale=0.3]{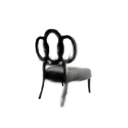} \\
                \includegraphics[scale=0.3]{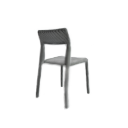} \\
                \includegraphics[scale=0.3]{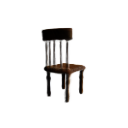} \\
                \includegraphics[scale=0.3]{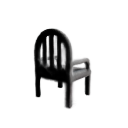} \\
                \includegraphics[scale=0.3]{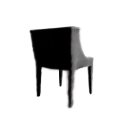} \\
                \includegraphics[scale=0.3]{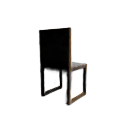} \\
                \includegraphics[scale=0.3]{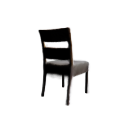} \\
                \includegraphics[scale=0.3]{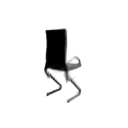} \\
                \includegraphics[scale=0.3]{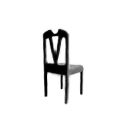} \\
                \includegraphics[scale=0.3]{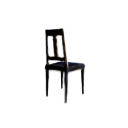}
            \end{minipage}
            \hspace{-1.5mm}
            \begin{minipage}{0.1\textwidth}
                \includegraphics[scale=0.3]{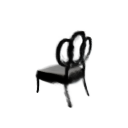} \\
                \includegraphics[scale=0.3]{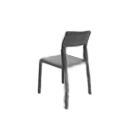} \\
                \includegraphics[scale=0.3]{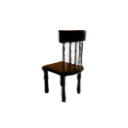} \\
                \includegraphics[scale=0.3]{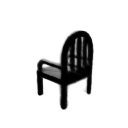} \\
                \includegraphics[scale=0.3]{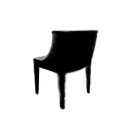} \\
                \includegraphics[scale=0.3]{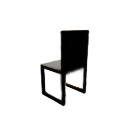} \\
                \includegraphics[scale=0.3]{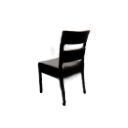} \\
                \includegraphics[scale=0.3]{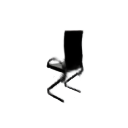} \\
                \includegraphics[scale=0.3]{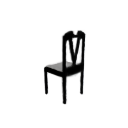} \\
                \includegraphics[scale=0.3]{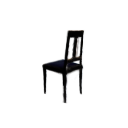}
            \end{minipage}
            \hspace{-1.5mm}
            \begin{minipage}{0.1\textwidth}
                \includegraphics[scale=0.3]{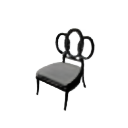} \\
                \includegraphics[scale=0.3]{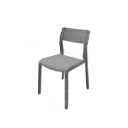} \\
                \includegraphics[scale=0.3]{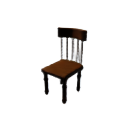} \\
                \includegraphics[scale=0.3]{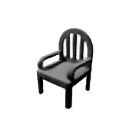} \\
                \includegraphics[scale=0.3]{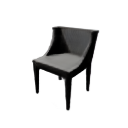} \\
                \includegraphics[scale=0.3]{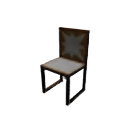} \\
                \includegraphics[scale=0.3]{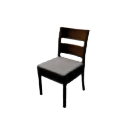} \\
                \includegraphics[scale=0.3]{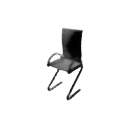} \\
                \includegraphics[scale=0.3]{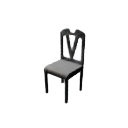} \\
                \includegraphics[scale=0.3]{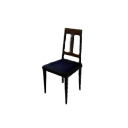}
            \end{minipage}
            \hspace{-1.5mm}
            \begin{minipage}{0.1\textwidth}
                \includegraphics[scale=0.3]{figures/qualitative/chairs/e53b07b648e8d041107a17cfae0b6df6/000080_pred_coarse.png} \\
                \includegraphics[scale=0.3]{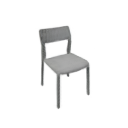} \\
                \includegraphics[scale=0.3]{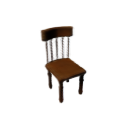} \\
                \includegraphics[scale=0.3]{figures/qualitative/chairs/ef3377832d90dbbacfe150564cb24aad/000080_pred_coarse.png} \\
                \includegraphics[scale=0.3]{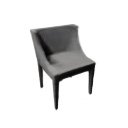} \\
                \includegraphics[scale=0.3]{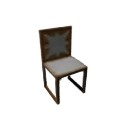} \\
                \includegraphics[scale=0.3]{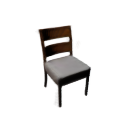} \\
                \includegraphics[scale=0.3]{figures/qualitative/chairs/9d395454d6de675d2025ebfdd95f4ba7/000080_pred_coarse.png} \\
                \includegraphics[scale=0.3]{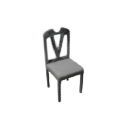} \\
                \includegraphics[scale=0.3]{figures/qualitative/chairs/c7e590c0390e8d5debe67d9b32c3ddf8/000080_pred_coarse.png}
            \end{minipage}
            \hspace{-1.5mm}
            \begin{minipage}{0.1\textwidth}
                \includegraphics[scale=0.3]{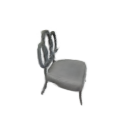} \\
                \includegraphics[scale=0.3]{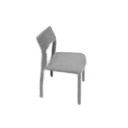} \\
                \includegraphics[scale=0.3]{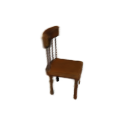} \\
                \includegraphics[scale=0.3]{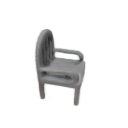} \\
                \includegraphics[scale=0.3]{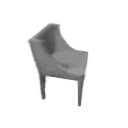} \\
                \includegraphics[scale=0.3]{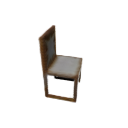} \\
                \includegraphics[scale=0.3]{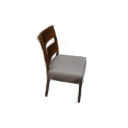} \\
                \includegraphics[scale=0.3]{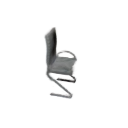} \\
                \includegraphics[scale=0.3]{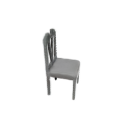} \\
                \includegraphics[scale=0.3]{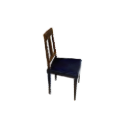}
            \end{minipage}
            \hspace{-1.5mm}
            \begin{minipage}{0.1\textwidth}
                \includegraphics[scale=0.3]{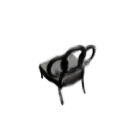} \\
                \includegraphics[scale=0.3]{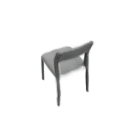} \\
                \includegraphics[scale=0.3]{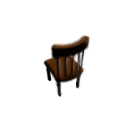} \\
                \includegraphics[scale=0.3]{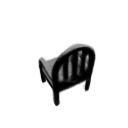} \\
                \includegraphics[scale=0.3]{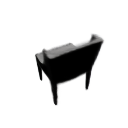} \\
                \includegraphics[scale=0.3]{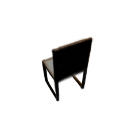} \\
                \includegraphics[scale=0.3]{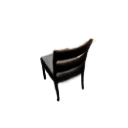} \\
                \includegraphics[scale=0.3]{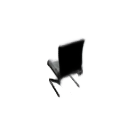} \\
                \includegraphics[scale=0.3]{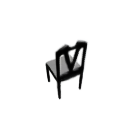} \\
                \includegraphics[scale=0.3]{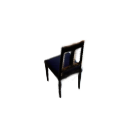}
            \end{minipage}
            \hspace{-1.5mm}
            \begin{minipage}{0.1\textwidth}
                \includegraphics[scale=0.3]{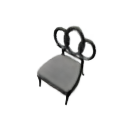} \\
                \includegraphics[scale=0.3]{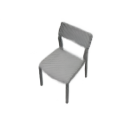} \\
                \includegraphics[scale=0.3]{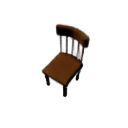} \\
                \includegraphics[scale=0.3]{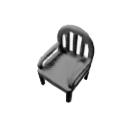} \\
                \includegraphics[scale=0.3]{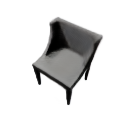} \\
                \includegraphics[scale=0.3]{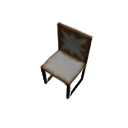} \\
                \includegraphics[scale=0.3]{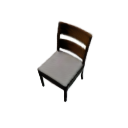} \\
                \includegraphics[scale=0.3]{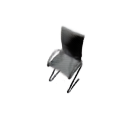} \\
                \includegraphics[scale=0.3]{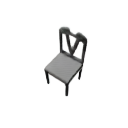} \\
                \includegraphics[scale=0.3]{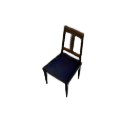}
            \end{minipage}
        \end{minipage}
    \end{minipage}
    \caption{Additional novel view synthesis results on the ``Chairs'' category of the ShapeNet-SRN~\cite{chang2015shapenet,sitzmann2019scene} dataset. As can be seen, ours can always synthesize photo-realistic and reasonable novel renderings from totally different viewpoints.}
    \label{fig:novel-view-synthesis-chairs}
\end{figure*}

\begin{figure*}
    \centering
    \begin{minipage}{\textwidth}
        \begin{minipage}{0.12\textwidth}
            \centering
            \subcaption*{Reference}
            \includegraphics[scale=0.3]{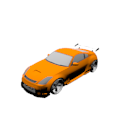} \\
            \includegraphics[scale=0.3]{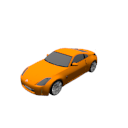} \\        \includegraphics[scale=0.3]{figures/qualitative/cars/90ba6416acd424e06d8db5f653b07b4b/src_rgb.png} \\
            \includegraphics[scale=0.3]{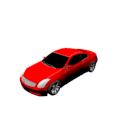} \\
            \includegraphics[scale=0.3]{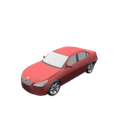} \\
            \includegraphics[scale=0.3]{figures/qualitative/cars/189c2b53ef76d02978a20fe14185667/src_rgb.png} \\
            \includegraphics[scale=0.3]{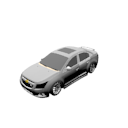} \\
            \includegraphics[scale=0.3]{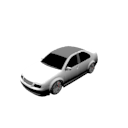} \\
            \includegraphics[scale=0.3]{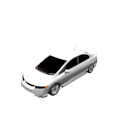} \\
            \includegraphics[scale=0.3]{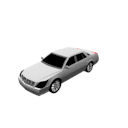}
        \end{minipage}
        \hspace{1mm}
        \vline
        \hspace{-3mm}
        \begin{minipage}{0.88\textwidth}
            \centering
            \subcaption*{Novel Views}
            \begin{minipage}{0.1\textwidth}
                \includegraphics[scale=0.3]{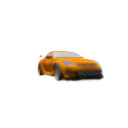} \\
                \includegraphics[scale=0.3]{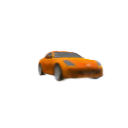} \\
                \includegraphics[scale=0.3]{figures/qualitative/cars/90ba6416acd424e06d8db5f653b07b4b/000000_pred_coarse.png} \\
                \includegraphics[scale=0.3]{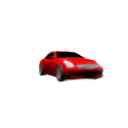} \\
                \includegraphics[scale=0.3]{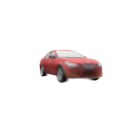} \\
                \includegraphics[scale=0.3]{figures/qualitative/cars/189c2b53ef76d02978a20fe14185667/000000_pred_coarse.png} \\
                \includegraphics[scale=0.3]{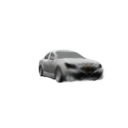} \\
                \includegraphics[scale=0.3]{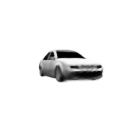} \\
                \includegraphics[scale=0.3]{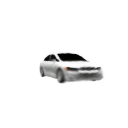} \\
                \includegraphics[scale=0.3]{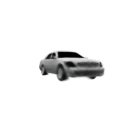}
            \end{minipage}
            \hspace{-1.5mm}
            \begin{minipage}{0.1\textwidth}
                \includegraphics[scale=0.3]{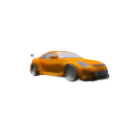} \\
                \includegraphics[scale=0.3]{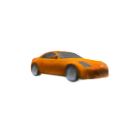} \\
                \includegraphics[scale=0.3]{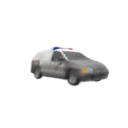} \\
                \includegraphics[scale=0.3]{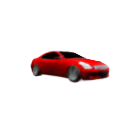} \\
                \includegraphics[scale=0.3]{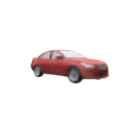} \\
                \includegraphics[scale=0.3]{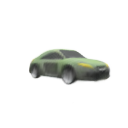} \\
                \includegraphics[scale=0.3]{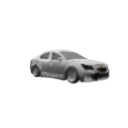} \\
                \includegraphics[scale=0.3]{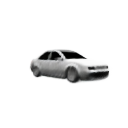} \\
                \includegraphics[scale=0.3]{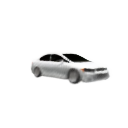} \\
                \includegraphics[scale=0.3]{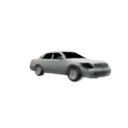}
            \end{minipage}
            \hspace{-1mm}
            \begin{minipage}{0.1\textwidth}
                \includegraphics[scale=0.3]{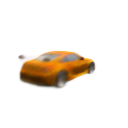} \\
                \includegraphics[scale=0.3]{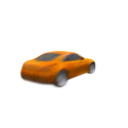} \\
                \includegraphics[scale=0.3]{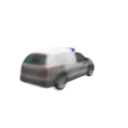} \\
                \includegraphics[scale=0.3]{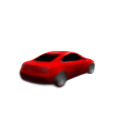} \\
                \includegraphics[scale=0.3]{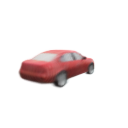} \\
                \includegraphics[scale=0.3]{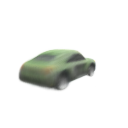} \\
                \includegraphics[scale=0.3]{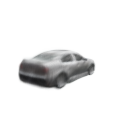} \\
                \includegraphics[scale=0.3]{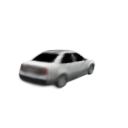} \\
                \includegraphics[scale=0.3]{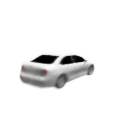} \\
                \includegraphics[scale=0.3]{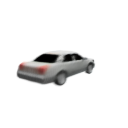}
            \end{minipage}
            \hspace{-2mm}
            \begin{minipage}{0.1\textwidth}
                \includegraphics[scale=0.3]{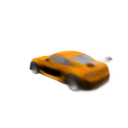} \\
                \includegraphics[scale=0.3]{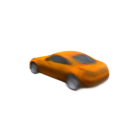} \\
                \includegraphics[scale=0.3]{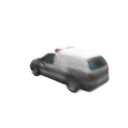} \\
                \includegraphics[scale=0.3]{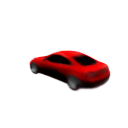} \\
                \includegraphics[scale=0.3]{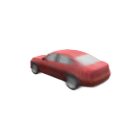} \\
                \includegraphics[scale=0.3]{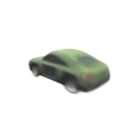} \\
                \includegraphics[scale=0.3]{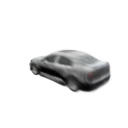} \\
                \includegraphics[scale=0.3]{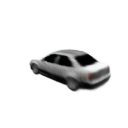} \\
                \includegraphics[scale=0.3]{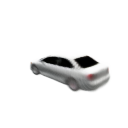} \\
                \includegraphics[scale=0.3]{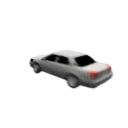}
            \end{minipage}
            \hspace{-1mm}
            \begin{minipage}{0.1\textwidth}
                \includegraphics[scale=0.3]{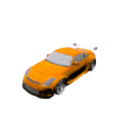} \\
                \includegraphics[scale=0.3]{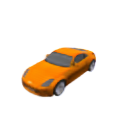} \\
                \includegraphics[scale=0.3]{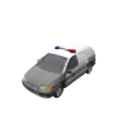} \\
                \includegraphics[scale=0.3]{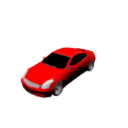} \\
                \includegraphics[scale=0.3]{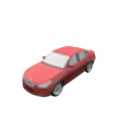} \\
                \includegraphics[scale=0.3]{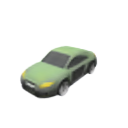} \\
                \includegraphics[scale=0.3]{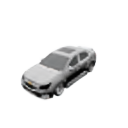} \\
                \includegraphics[scale=0.3]{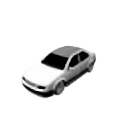} \\
                \includegraphics[scale=0.3]{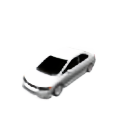} \\
                \includegraphics[scale=0.3]{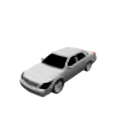}
            \end{minipage}
            \hspace{-2.3mm}
            \begin{minipage}{0.1\textwidth}
                \includegraphics[scale=0.3]{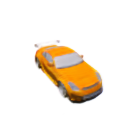} \\
                \includegraphics[scale=0.3]{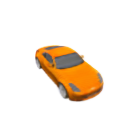} \\
                \includegraphics[scale=0.3]{figures/qualitative/cars/90ba6416acd424e06d8db5f653b07b4b/000080_pred_coarse.png} \\
                \includegraphics[scale=0.3]{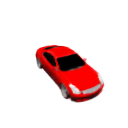} \\
                \includegraphics[scale=0.3]{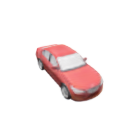} \\
                \includegraphics[scale=0.3]{figures/qualitative/cars/189c2b53ef76d02978a20fe14185667/000080_pred_coarse.png} \\
                \includegraphics[scale=0.3]{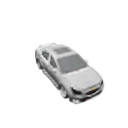} \\
                \includegraphics[scale=0.3]{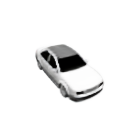} \\
                \includegraphics[scale=0.3]{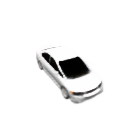} \\
                \includegraphics[scale=0.3]{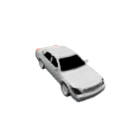}
            \end{minipage}
            \hspace{-1.5mm}
            \begin{minipage}{0.1\textwidth}
                \includegraphics[scale=0.3]{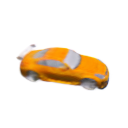} \\
                \includegraphics[scale=0.3]{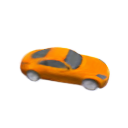} \\
                \includegraphics[scale=0.3]{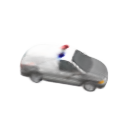} \\
                \includegraphics[scale=0.3]{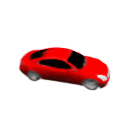} \\
                \includegraphics[scale=0.3]{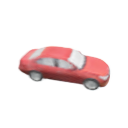} \\
                \includegraphics[scale=0.3]{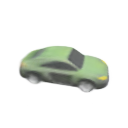} \\
                \includegraphics[scale=0.3]{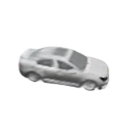} \\
                \includegraphics[scale=0.3]{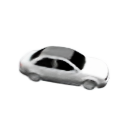} \\
                \includegraphics[scale=0.3]{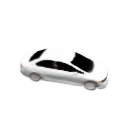} \\
                \includegraphics[scale=0.3]{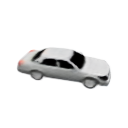}
            \end{minipage}
            \hspace{-1mm}
            \begin{minipage}{0.1\textwidth}
                \includegraphics[scale=0.3]{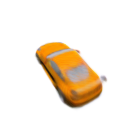} \\
                \includegraphics[scale=0.3]{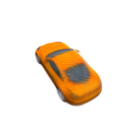} \\
                \includegraphics[scale=0.3]{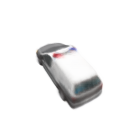} \\
                \includegraphics[scale=0.3]{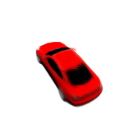} \\
                \includegraphics[scale=0.3]{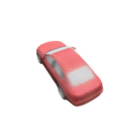} \\
                \includegraphics[scale=0.3]{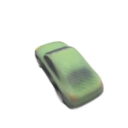} \\
                \includegraphics[scale=0.3]{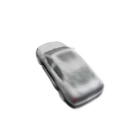} \\
                \includegraphics[scale=0.3]{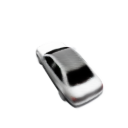} \\
                \includegraphics[scale=0.3]{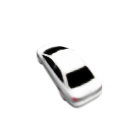} \\
                \includegraphics[scale=0.3]{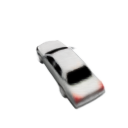}
            \end{minipage}
            \hspace{-1.5mm}
            \begin{minipage}{0.1\textwidth}
                \includegraphics[scale=0.3]{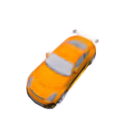} \\
                \includegraphics[scale=0.3]{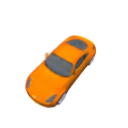} \\
                \includegraphics[scale=0.3]{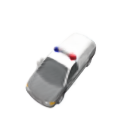} \\
                \includegraphics[scale=0.3]{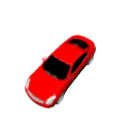} \\
                \includegraphics[scale=0.3]{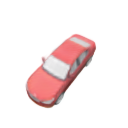} \\
                \includegraphics[scale=0.3]{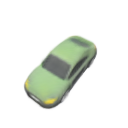} \\
                \includegraphics[scale=0.3]{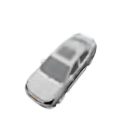} \\
                \includegraphics[scale=0.3]{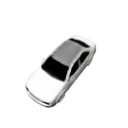} \\
                \includegraphics[scale=0.3]{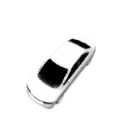} \\
                \includegraphics[scale=0.3]{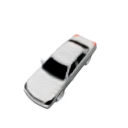}
            \end{minipage}
        \end{minipage}
    \end{minipage}
    \caption{Additional novel view synthesis results on the ``Cars'' category of the ShapeNet-SRN~\cite{chang2015shapenet,sitzmann2019scene} dataset. As can be seen, ours can always synthesize photo-realistic and reasonable novel renderings from totally different viewpoints.}
    \label{fig:novel-view-synthesis-cars}
\end{figure*}

We further provide more visualization of novel view synthesis results in Fig.~\ref{fig:novel-view-synthesis-chairs} and Fig.~\ref{fig:novel-view-synthesis-cars}. As can be seen, SymmNeRF can always synthesize photo-realistic and reasonable novel renderings from totally different viewpoints. With the help of the symmetry priors and the hypernetwork, SymmNeRF accurately recovers the geometry information and texture details despite the occlusions in the reference view.

\subsection{Generalization Results on Real-World Datasets}
\begin{figure*}
    \centering
    \begin{minipage}{\textwidth}
        \begin{minipage}{0.12\textwidth}
            \centering
            \subcaption*{Reference}
            \includegraphics[scale=0.3]{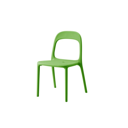} \\
            \includegraphics[scale=0.3]{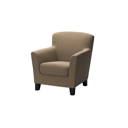} \\
            \includegraphics[scale=0.3]{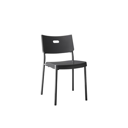} \\
            \includegraphics[scale=0.3]{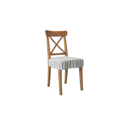} \\
            \includegraphics[scale=0.3]{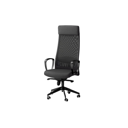} \\
            \includegraphics[scale=0.3]{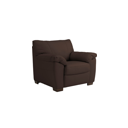} \\
            \includegraphics[scale=0.3]{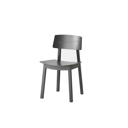} \\
            \includegraphics[scale=0.3]{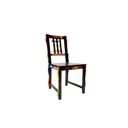} \\
            \includegraphics[scale=0.3]{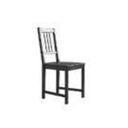} \\
            \includegraphics[scale=0.3]{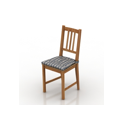}
        \end{minipage}
        \begin{minipage}{0.88\textwidth}
            \begin{minipage}{0.5\textwidth}
                \centering
                \subcaption*{PixelNeRF}
                \begin{minipage}{0.25\textwidth}
                    \includegraphics[scale=0.3]{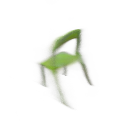} \\
                    \includegraphics[scale=0.3]{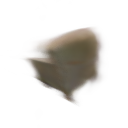} \\
                    \includegraphics[scale=0.3]{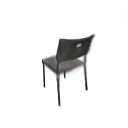} \\
                    \includegraphics[scale=0.3]{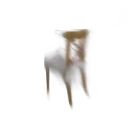} \\
                    \includegraphics[scale=0.3]{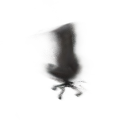} \\
                    \includegraphics[scale=0.3]{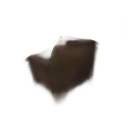} \\
                    \includegraphics[scale=0.3]{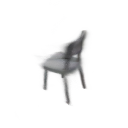} \\
                    \includegraphics[scale=0.3]{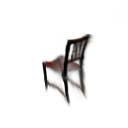} \\
                    \includegraphics[scale=0.3]{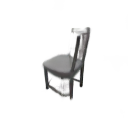} \\
                    \includegraphics[scale=0.3]{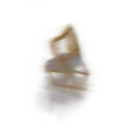}
                \end{minipage}
                \hspace{-3mm}
                \begin{minipage}{0.25\textwidth}
                    \includegraphics[scale=0.3]{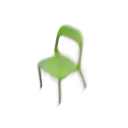} \\
                    \includegraphics[scale=0.3]{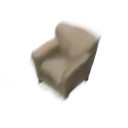} \\
                    \includegraphics[scale=0.3]{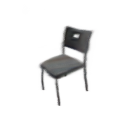} \\
                    \includegraphics[scale=0.3]{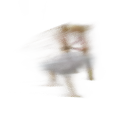} \\
                    \includegraphics[scale=0.3]{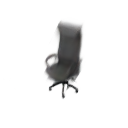} \\
                    \includegraphics[scale=0.3]{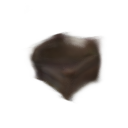} \\
                    \includegraphics[scale=0.3]{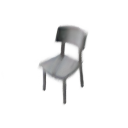} \\
                    \includegraphics[scale=0.3]{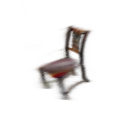} \\
                    \includegraphics[scale=0.3]{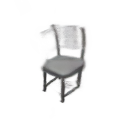} \\
                    \includegraphics[scale=0.3]{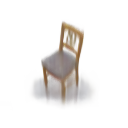}
                \end{minipage}
                \hspace{-3mm}
                \begin{minipage}{0.25\textwidth}
                    \includegraphics[scale=0.3]{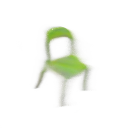} \\
                    \includegraphics[scale=0.3]{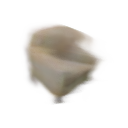} \\
                    \includegraphics[scale=0.3]{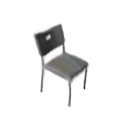} \\
                    \includegraphics[scale=0.3]{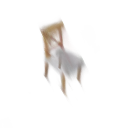} \\
                    \includegraphics[scale=0.3]{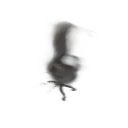} \\
                    \includegraphics[scale=0.3]{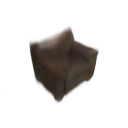} \\
                    \includegraphics[scale=0.3]{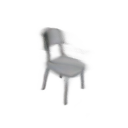} \\
                    \includegraphics[scale=0.3]{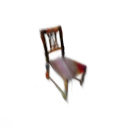} \\
                    \includegraphics[scale=0.3]{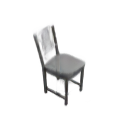} \\
                    \includegraphics[scale=0.3]{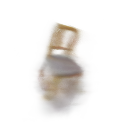}
                \end{minipage}
                \hspace{-3mm}
                \begin{minipage}{0.25\textwidth}
                    \includegraphics[scale=0.3]{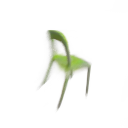} \\
                    \includegraphics[scale=0.3]{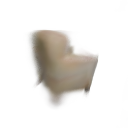} \\
                    \includegraphics[scale=0.3]{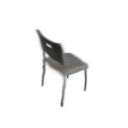} \\
                    \includegraphics[scale=0.3]{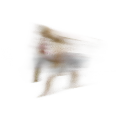} \\
                    \includegraphics[scale=0.3]{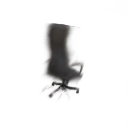} \\
                    \includegraphics[scale=0.3]{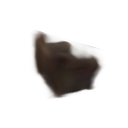} \\
                    \includegraphics[scale=0.3]{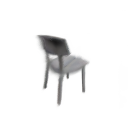} \\
                    \includegraphics[scale=0.3]{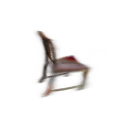} \\
                    \includegraphics[scale=0.3]{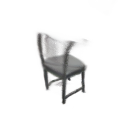} \\
                    \includegraphics[scale=0.3]{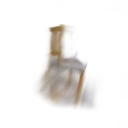}
                \end{minipage}
            \end{minipage}
            \hspace{-3mm}
            \begin{minipage}{0.5\textwidth}
                \centering
                \subcaption*{Ours}
                \begin{minipage}{0.25\textwidth}
                    \includegraphics[scale=0.3]{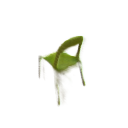} \\
                    \includegraphics[scale=0.3]{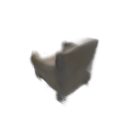} \\
                    \includegraphics[scale=0.3]{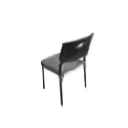} \\
                    \includegraphics[scale=0.3]{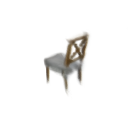} \\
                    \includegraphics[scale=0.3]{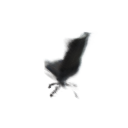} \\
                    \includegraphics[scale=0.3]{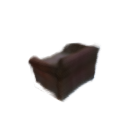} \\
                    \includegraphics[scale=0.3]{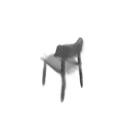} \\
                    \includegraphics[scale=0.3]{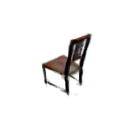} \\
                    \includegraphics[scale=0.3]{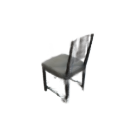} \\
                    \includegraphics[scale=0.3]{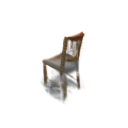}
                \end{minipage}
                \hspace{-3mm}
                \begin{minipage}{0.25\textwidth}
                    \includegraphics[scale=0.3]{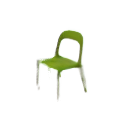} \\
                    \includegraphics[scale=0.3]{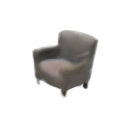} \\
                    \includegraphics[scale=0.3]{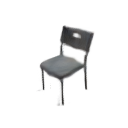} \\
                    \includegraphics[scale=0.3]{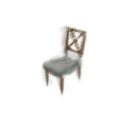} \\
                    \includegraphics[scale=0.3]{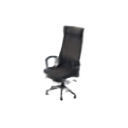} \\
                    \includegraphics[scale=0.3]{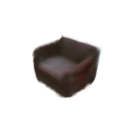} \\
                    \includegraphics[scale=0.3]{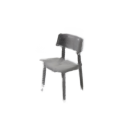} \\
                    \includegraphics[scale=0.3]{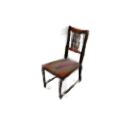} \\
                    \includegraphics[scale=0.3]{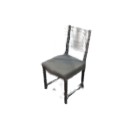} \\
                    \includegraphics[scale=0.3]{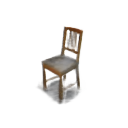}
                \end{minipage}
                \hspace{-3mm}
                \begin{minipage}{0.25\textwidth}
                    \includegraphics[scale=0.3]{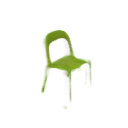} \\
                    \includegraphics[scale=0.3]{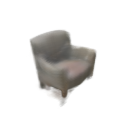} \\
                    \includegraphics[scale=0.3]{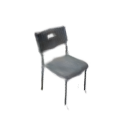} \\
                    \includegraphics[scale=0.3]{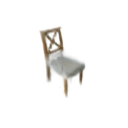} \\
                    \includegraphics[scale=0.3]{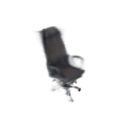} \\
                    \includegraphics[scale=0.3]{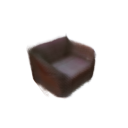} \\
                    \includegraphics[scale=0.3]{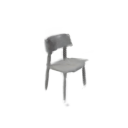} \\
                    \includegraphics[scale=0.3]{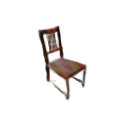} \\
                    \includegraphics[scale=0.3]{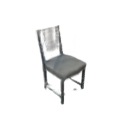} \\
                    \includegraphics[scale=0.3]{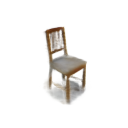}
                \end{minipage}
                \hspace{-3mm}
                \begin{minipage}{0.25\textwidth}
                    \includegraphics[scale=0.3]{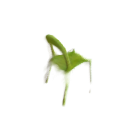} \\
                    \includegraphics[scale=0.3]{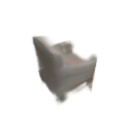} \\
                    \includegraphics[scale=0.3]{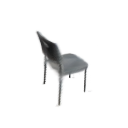} \\
                    \includegraphics[scale=0.3]{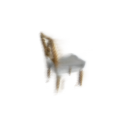} \\
                    \includegraphics[scale=0.3]{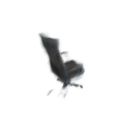} \\
                    \includegraphics[scale=0.3]{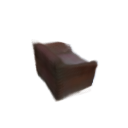} \\
                    \includegraphics[scale=0.3]{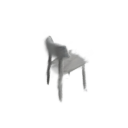} \\
                    \includegraphics[scale=0.3]{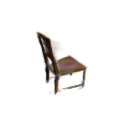} \\
                    \includegraphics[scale=0.3]{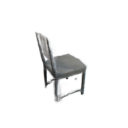} \\
                    \includegraphics[scale=0.3]{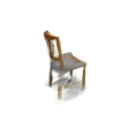}
                \end{minipage}
            \end{minipage}
        \end{minipage}
    \end{minipage}
    \caption{Additional qualitative comparisons with PixelNeRF~\cite{yu2021pixelnerf} on the real-world Pix3D~\cite{sun2018pix3d} dataset. Compared with PixelNeRF, SymmNeRF yields better generalization.} 
    \label{fig:real-world-chairs}
\end{figure*}

\begin{figure*}
    \centering
    \begin{minipage}{\textwidth}
        \begin{minipage}{0.12\textwidth}
            \centering
            \subcaption*{Reference}
            \includegraphics[scale=0.3]{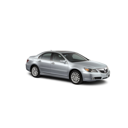} \\
            \includegraphics[scale=0.3]{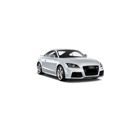} \\
            \includegraphics[scale=0.3]{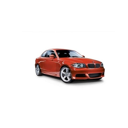} \\
            \includegraphics[scale=0.3]{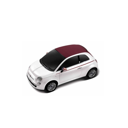} \\
            \includegraphics[scale=0.3]{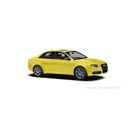} \\
            \includegraphics[scale=0.3]{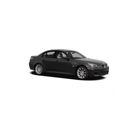} \\
            \includegraphics[scale=0.3]{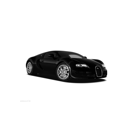} \\
            \includegraphics[scale=0.3]{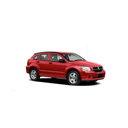} \\
            \includegraphics[scale=0.3]{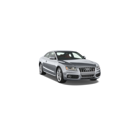} \\
            \includegraphics[scale=0.3]{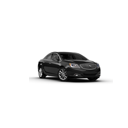}
        \end{minipage}
        \begin{minipage}{0.88\textwidth}
            \begin{minipage}{0.5\textwidth}
                \centering
                \subcaption*{PixelNeRF}
                \begin{minipage}{0.25\textwidth}
                    \includegraphics[scale=0.3]{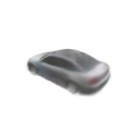} \\
                    \includegraphics[scale=0.3]{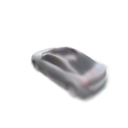} \\
                    \includegraphics[scale=0.3]{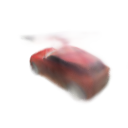} \\
                    \includegraphics[scale=0.3]{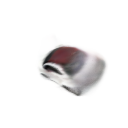} \\
                    \includegraphics[scale=0.3]{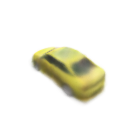} \\
                    \includegraphics[scale=0.3]{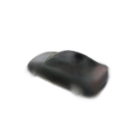} \\
                    \includegraphics[scale=0.3]{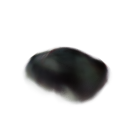} \\
                    \includegraphics[scale=0.3]{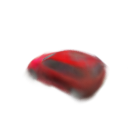} \\
                    \includegraphics[scale=0.3]{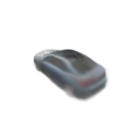} \\
                    \includegraphics[scale=0.3]{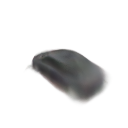}
                \end{minipage}
                \hspace{-2mm}
                \begin{minipage}{0.25\textwidth}
                    \includegraphics[scale=0.3]{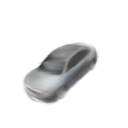} \\
                    \includegraphics[scale=0.3]{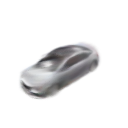} \\
                    \includegraphics[scale=0.3]{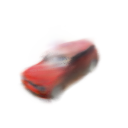} \\
                    \includegraphics[scale=0.3]{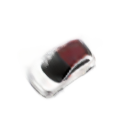} \\
                    \includegraphics[scale=0.3]{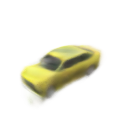} \\
                    \includegraphics[scale=0.3]{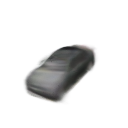} \\
                    \includegraphics[scale=0.3]{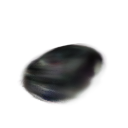} \\
                    \includegraphics[scale=0.3]{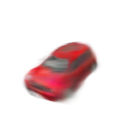} \\
                    \includegraphics[scale=0.3]{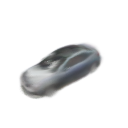} \\
                    \includegraphics[scale=0.3]{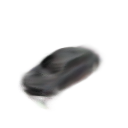}
                \end{minipage}
                \hspace{-2mm}
                \begin{minipage}{0.25\textwidth}
                    \includegraphics[scale=0.3]{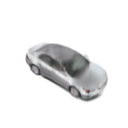} \\
                    \includegraphics[scale=0.3]{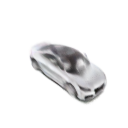} \\
                    \includegraphics[scale=0.3]{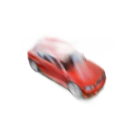} \\
                    \includegraphics[scale=0.3]{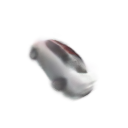} \\
                    \includegraphics[scale=0.3]{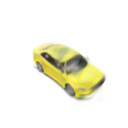} \\
                    \includegraphics[scale=0.3]{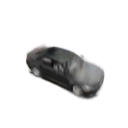} \\
                    \includegraphics[scale=0.3]{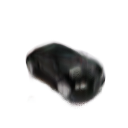} \\
                    \includegraphics[scale=0.3]{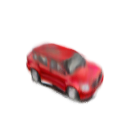} \\
                    \includegraphics[scale=0.3]{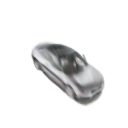} \\
                    \includegraphics[scale=0.3]{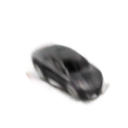}
                \end{minipage}
                \hspace{-2mm}
                \begin{minipage}{0.25\textwidth}
                    \includegraphics[scale=0.3]{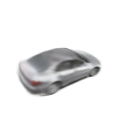} \\
                    \includegraphics[scale=0.3]{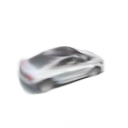} \\
                    \includegraphics[scale=0.3]{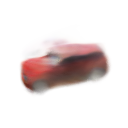} \\
                    \includegraphics[scale=0.3]{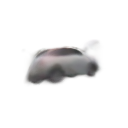} \\
                    \includegraphics[scale=0.3]{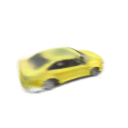} \\
                    \includegraphics[scale=0.3]{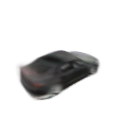} \\
                    \includegraphics[scale=0.3]{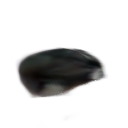} \\
                    \includegraphics[scale=0.3]{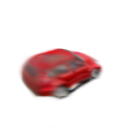} \\
                    \includegraphics[scale=0.3]{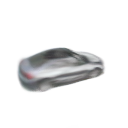} \\
                    \includegraphics[scale=0.3]{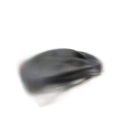}
                \end{minipage}
            \end{minipage}
            \hspace{-3mm}
            \begin{minipage}{0.5\textwidth}
                \centering
                \subcaption*{Ours}
                \begin{minipage}{0.25\textwidth}
                    \includegraphics[scale=0.3]{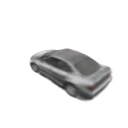} \\
                    \includegraphics[scale=0.3]{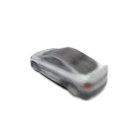} \\
                    \includegraphics[scale=0.3]{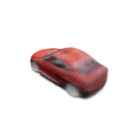} \\
                    \includegraphics[scale=0.3]{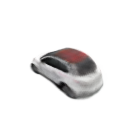} \\
                    \includegraphics[scale=0.3]{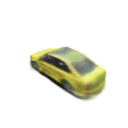} \\
                    \includegraphics[scale=0.3]{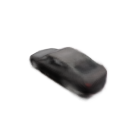} \\
                    \includegraphics[scale=0.3]{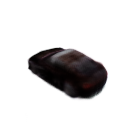} \\
                    \includegraphics[scale=0.3]{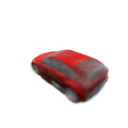} \\
                    \includegraphics[scale=0.3]{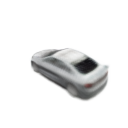} \\
                    \includegraphics[scale=0.3]{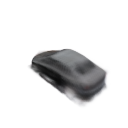}
                \end{minipage}
                \hspace{-2mm}
                \begin{minipage}{0.25\textwidth}
                    \includegraphics[scale=0.3]{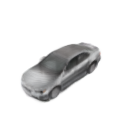} \\
                    \includegraphics[scale=0.3]{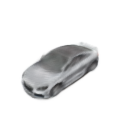} \\
                    \includegraphics[scale=0.3]{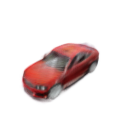} \\
                    \includegraphics[scale=0.3]{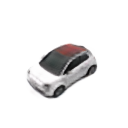} \\
                    \includegraphics[scale=0.3]{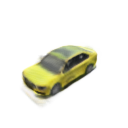} \\
                    \includegraphics[scale=0.3]{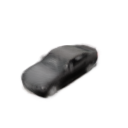} \\
                    \includegraphics[scale=0.3]{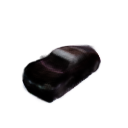} \\
                    \includegraphics[scale=0.3]{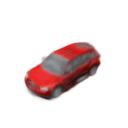} \\
                    \includegraphics[scale=0.3]{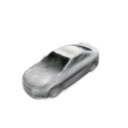} \\
                    \includegraphics[scale=0.3]{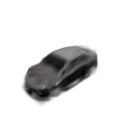}
                \end{minipage}
                \hspace{-2mm}
                \begin{minipage}{0.25\textwidth}
                    \includegraphics[scale=0.3]{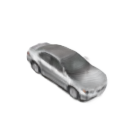} \\
                    \includegraphics[scale=0.3]{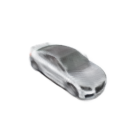} \\
                    \includegraphics[scale=0.3]{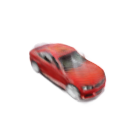} \\
                    \includegraphics[scale=0.3]{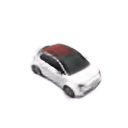} \\
                    \includegraphics[scale=0.3]{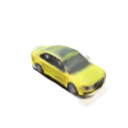} \\
                    \includegraphics[scale=0.3]{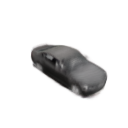} \\
                    \includegraphics[scale=0.3]{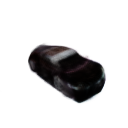} \\
                    \includegraphics[scale=0.3]{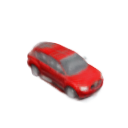} \\
                    \includegraphics[scale=0.3]{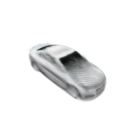} \\
                    \includegraphics[scale=0.3]{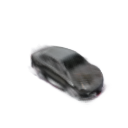}
                \end{minipage}
                \hspace{-2mm}
                \begin{minipage}{0.25\textwidth}
                    \includegraphics[scale=0.3]{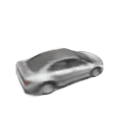} \\
                    \includegraphics[scale=0.3]{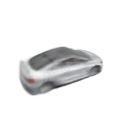} \\
                    \includegraphics[scale=0.3]{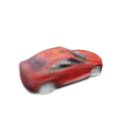} \\
                    \includegraphics[scale=0.3]{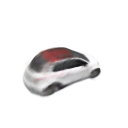} \\
                    \includegraphics[scale=0.3]{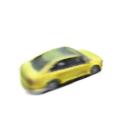} \\
                    \includegraphics[scale=0.3]{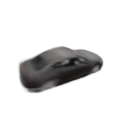} \\
                    \includegraphics[scale=0.3]{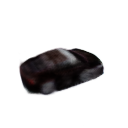} \\
                    \includegraphics[scale=0.3]{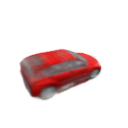} \\
                    \includegraphics[scale=0.3]{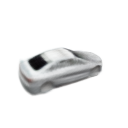} \\
                    \includegraphics[scale=0.3]{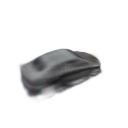}
                \end{minipage}
            \end{minipage}
        \end{minipage}
    \end{minipage}
    \caption{Additional qualitative comparisons with PixelNeRF~\cite{yu2021pixelnerf} on the real-world Stanford Cars~\cite{krause20133d} dataset. Compared with PixelNeRF, SymmNeRF yields better generalization.} 
    \label{fig:real-world-cars}
\end{figure*}

Here we provide additional generalization results on the real-world Pix3D~\cite{sun2018pix3d} and Stanford Cars~\cite{krause20133d} datasets in Fig.~\ref{fig:real-world-chairs} and Fig.~\ref{fig:real-world-cars}. Compared with PixelNeRF~\cite{yu2021pixelnerf}, SymmNeRF can effectively infer the geometry and appearance of real-world chairs and cars.

\subsection{Qualitative comparisons on the ShapeNet-NMR dataset under the category-agnostic single-view reconstruction setting}
\begin{figure*}[t]
    \centering
    \includegraphics[scale=0.27]{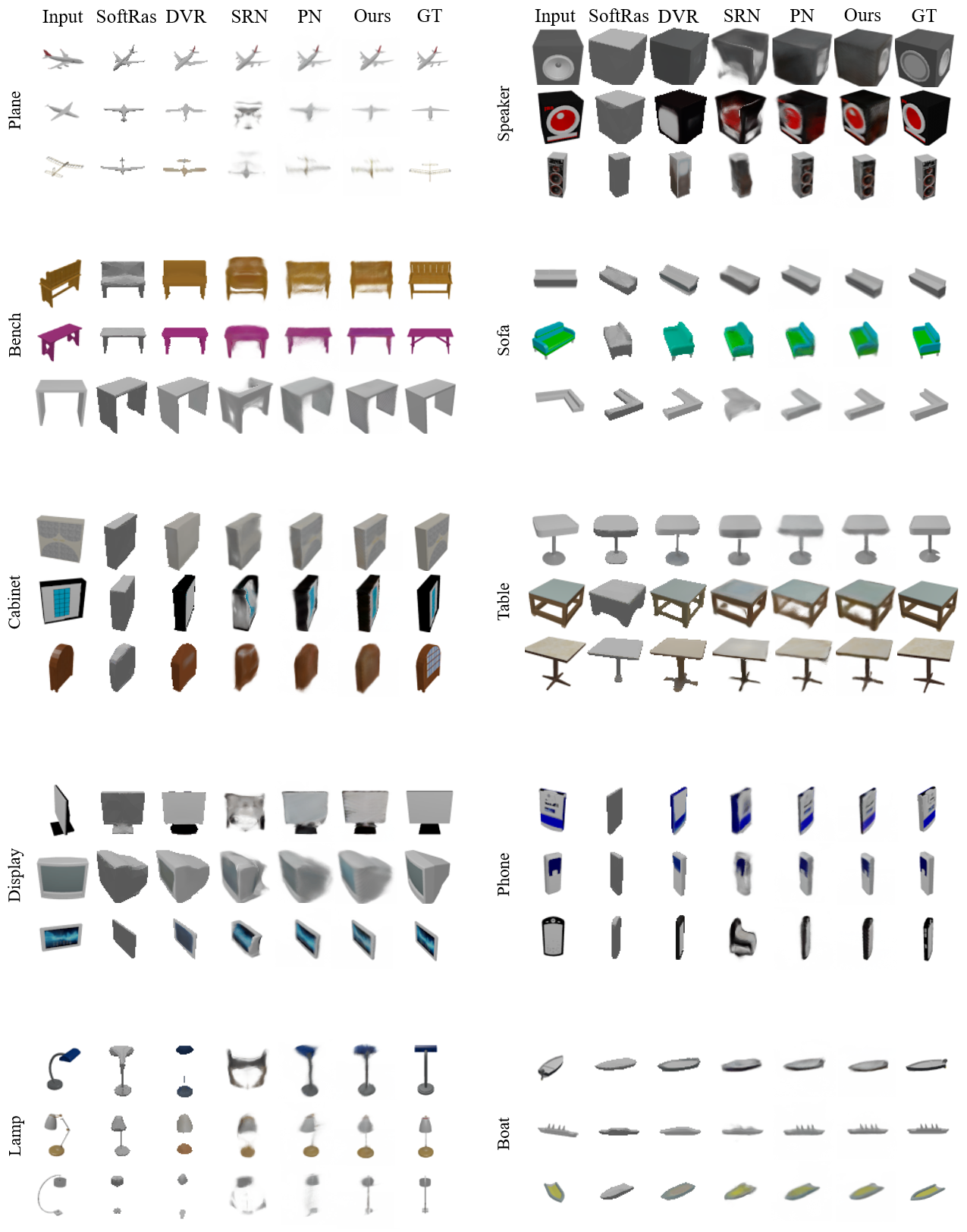}
    \caption{Addtional qualitative comparisons on the ShapeNet-NMR~\cite{chang2015shapenet,kato2018renderer} dataset under the category-agnostic single-view reconstruction setting.}
    \label{fig:nmr-more}
\end{figure*}

\begin{figure*}[t]
    \centering
    \includegraphics[scale=0.26]{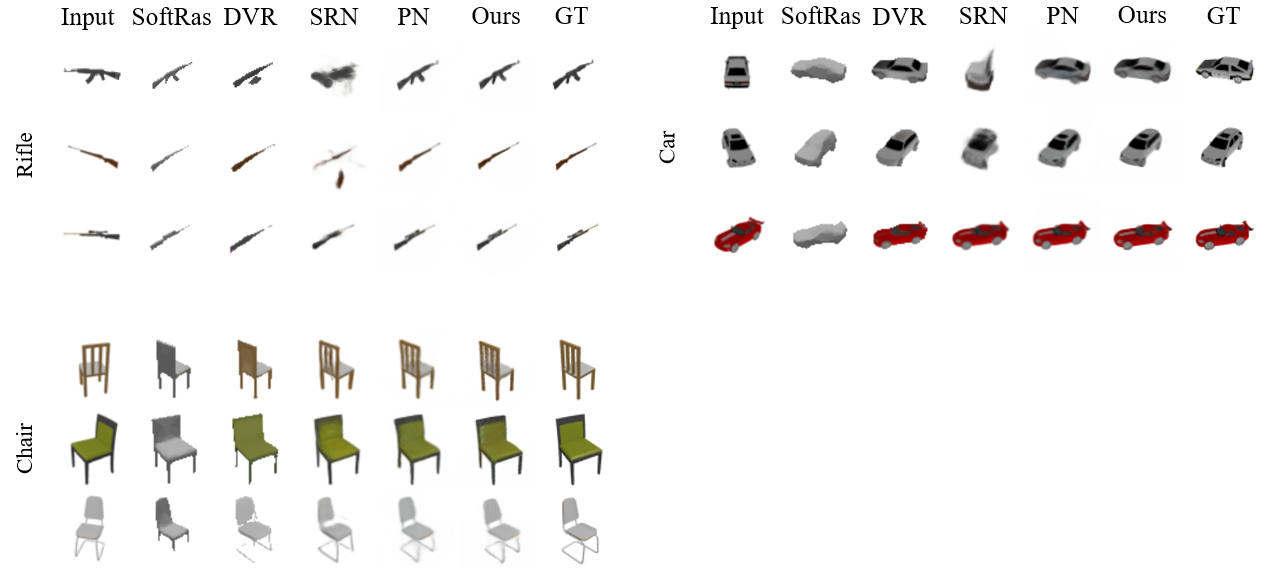}
    \caption{Addtional qualitative comparisons on the ShapeNet-NMR~\cite{chang2015shapenet,kato2018renderer} dataset under the category-agnostic single-view reconstruction setting.}
    \label{fig:nmr-more-part2}
\end{figure*}

We provide additional qualitative comparisons on the ShapeNet-NMR~\cite{chang2015shapenet,kato2018renderer} dataset under the category-agnostic single-view reconstruction setting. We show in Fig.~\ref{fig:nmr-more} and Fig.~\ref{fig:nmr-more-part2} that SymmNeRF outperforms other state-of-the-art methods~\cite{liu2019softras,niemeyer2020differentiable,sitzmann2019scene,yu2021pixelnerf}. This also implies that symmetry priors benefit the reconstruction of almost all symmetric objects, and that our method can also deal with objects that are not perfectly symmetric. This is because a few asymmetric objects are also included in the training dataset. Our model can perceive and recognize asymmetry thanks to the global latent code and hypernetwork. SymmNeRF therefore adaptively chooses to utilize local features to reconstruct asymmetric objects.

\subsection{Ablation Study}
\begin{figure*}
    \centering
    \begin{minipage}{\textwidth}
        \begin{minipage}{0.12\textwidth}
            \centering
            \subcaption*{Reference}
            \includegraphics[scale=0.25]{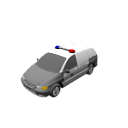} \\
            \includegraphics[scale=0.25]{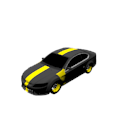} \\
            \includegraphics[scale=0.25]{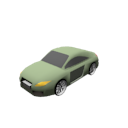} \\
            \includegraphics[scale=0.25]{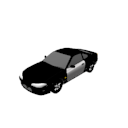} \\
            \includegraphics[scale=0.25]{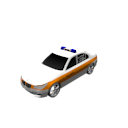} \\
            \includegraphics[scale=0.25]{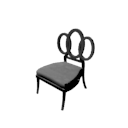} \\
            \includegraphics[scale=0.25]{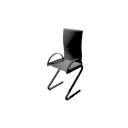} \\
            \includegraphics[scale=0.25]{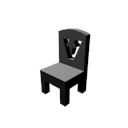} \\
            \includegraphics[scale=0.25]{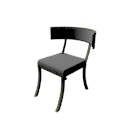} \\
            \includegraphics[scale=0.25]{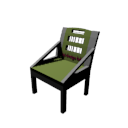}
        \end{minipage}
        \begin{minipage}{0.88\textwidth}
            \begin{minipage}{0.20\textwidth}
                \centering
                \subcaption*{(a)}
                \begin{minipage}{0.5\textwidth}
                    \includegraphics[scale=0.25]{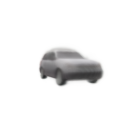} \\
                    \includegraphics[scale=0.25]{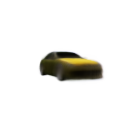} \\
                    \includegraphics[scale=0.25]{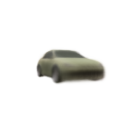} \\
                    \includegraphics[scale=0.25]{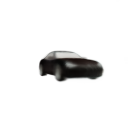} \\
                    \includegraphics[scale=0.25]{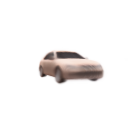} \\
                    \includegraphics[scale=0.25]{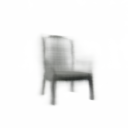} \\
                    \includegraphics[scale=0.25]{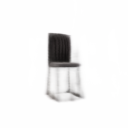} \\
                    \includegraphics[scale=0.25]{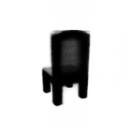} \\
                    \includegraphics[scale=0.25]{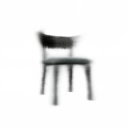} \\
                    \includegraphics[scale=0.25]{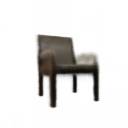}
                \end{minipage}
                \hspace{-3mm}
                \begin{minipage}{0.5\textwidth}
                    \includegraphics[scale=0.25]{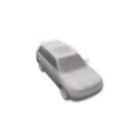} \\
                    \includegraphics[scale=0.25]{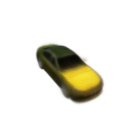} \\
                    \includegraphics[scale=0.25]{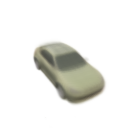} \\
                    \includegraphics[scale=0.25]{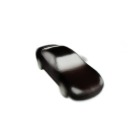} \\
                    \includegraphics[scale=0.25]{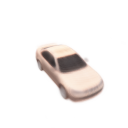} \\
                    \includegraphics[scale=0.25]{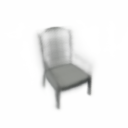} \\
                    \includegraphics[scale=0.25]{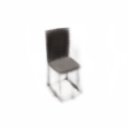} \\
                    \includegraphics[scale=0.25]{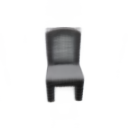} \\
                    \includegraphics[scale=0.25]{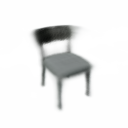} \\
                    \includegraphics[scale=0.25]{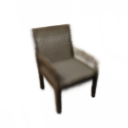}
                \end{minipage}
            \end{minipage}
            \hspace{-3mm}
            \begin{minipage}{0.20\textwidth}
                \centering
                \subcaption*{(b)}
                \begin{minipage}{0.5\textwidth}
                    \includegraphics[scale=0.25]{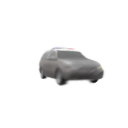} \\
                    \includegraphics[scale=0.25]{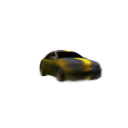} \\
                    \includegraphics[scale=0.25]{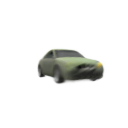} \\
                    \includegraphics[scale=0.25]{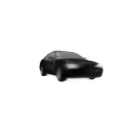} \\
                    \includegraphics[scale=0.25]{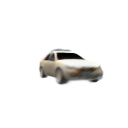} \\
                    \includegraphics[scale=0.25]{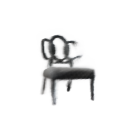} \\
                    \includegraphics[scale=0.25]{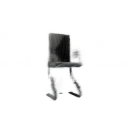} \\
                    \includegraphics[scale=0.25]{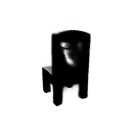} \\
                    \includegraphics[scale=0.25]{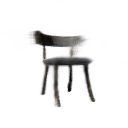} \\
                    \includegraphics[scale=0.25]{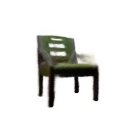}
                \end{minipage}
                \hspace{-3mm}
                \begin{minipage}{0.5\textwidth}
                    \includegraphics[scale=0.25]{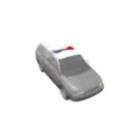} \\
                    \includegraphics[scale=0.25]{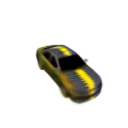} \\
                    \includegraphics[scale=0.25]{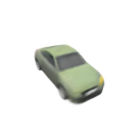} \\
                    \includegraphics[scale=0.25]{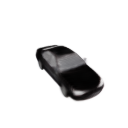} \\
                    \includegraphics[scale=0.25]{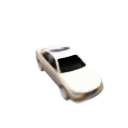} \\
                    \includegraphics[scale=0.25]{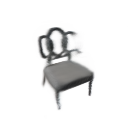} \\
                    \includegraphics[scale=0.25]{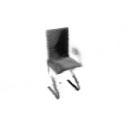} \\
                    \includegraphics[scale=0.25]{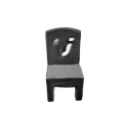} \\
                    \includegraphics[scale=0.25]{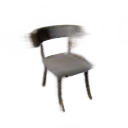} \\
                    \includegraphics[scale=0.25]{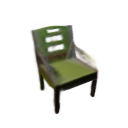}
                \end{minipage}
            \end{minipage}
            \hspace{-3mm}
            \begin{minipage}{0.20\textwidth}
                \centering
                \subcaption*{(c)}
                \begin{minipage}{0.5\textwidth}
                    \includegraphics[scale=0.25]{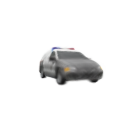} \\
                    \includegraphics[scale=0.25]{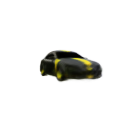} \\
                    \includegraphics[scale=0.25]{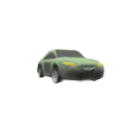} \\
                    \includegraphics[scale=0.25]{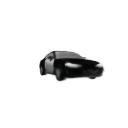} \\
                    \includegraphics[scale=0.25]{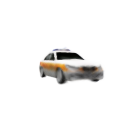} \\
                    \includegraphics[scale=0.25]{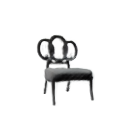} \\
                    \includegraphics[scale=0.25]{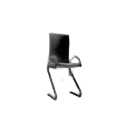} \\
                    \includegraphics[scale=0.25]{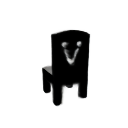} \\
                    \includegraphics[scale=0.25]{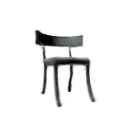} \\
                    \includegraphics[scale=0.25]{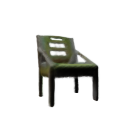}
                \end{minipage}
                \hspace{-3mm}
                \begin{minipage}{0.5\textwidth}
                    \includegraphics[scale=0.25]{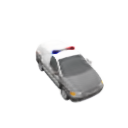} \\
                    \includegraphics[scale=0.25]{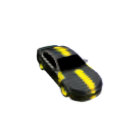} \\
                    \includegraphics[scale=0.25]{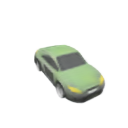} \\
                    \includegraphics[scale=0.25]{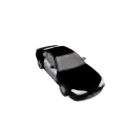} \\
                    \includegraphics[scale=0.25]{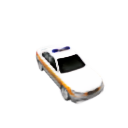} \\
                    \includegraphics[scale=0.25]{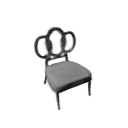} \\
                    \includegraphics[scale=0.25]{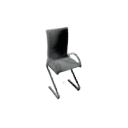} \\
                    \includegraphics[scale=0.25]{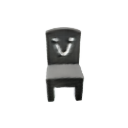} \\
                    \includegraphics[scale=0.25]{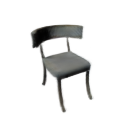} \\
                    \includegraphics[scale=0.25]{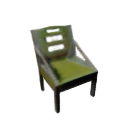}
                \end{minipage}
            \end{minipage}
            \hspace{-3mm}
            \begin{minipage}{0.20\textwidth}
                \centering
                \subcaption*{(d)}
                \begin{minipage}{0.5\textwidth}
                    \includegraphics[scale=0.25]{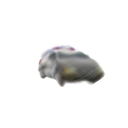} \\
                    \includegraphics[scale=0.25]{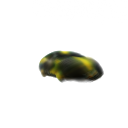} \\
                    \includegraphics[scale=0.25]{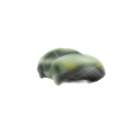} \\
                    \includegraphics[scale=0.25]{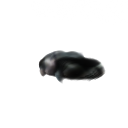} \\
                    \includegraphics[scale=0.25]{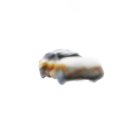} \\
                    \includegraphics[scale=0.25]{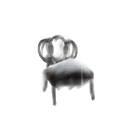} \\
                    \includegraphics[scale=0.25]{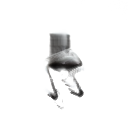} \\
                    \includegraphics[scale=0.25]{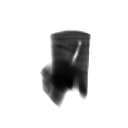} \\
                    \includegraphics[scale=0.25]{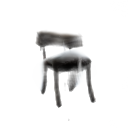} \\
                    \includegraphics[scale=0.25]{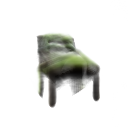}
                \end{minipage}
                \hspace{-3mm}
                \begin{minipage}{0.5\textwidth}
                    \includegraphics[scale=0.25]{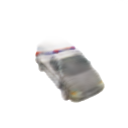} \\
                    \includegraphics[scale=0.25]{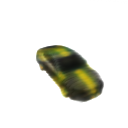} \\
                    \includegraphics[scale=0.25]{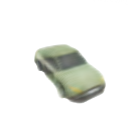} \\
                    \includegraphics[scale=0.25]{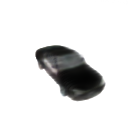} \\
                    \includegraphics[scale=0.25]{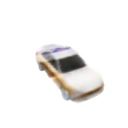} \\
                    \includegraphics[scale=0.25]{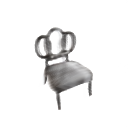} \\
                    \includegraphics[scale=0.25]{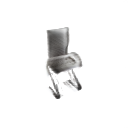} \\
                    \includegraphics[scale=0.25]{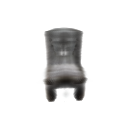} \\
                    \includegraphics[scale=0.25]{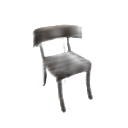} \\
                    \includegraphics[scale=0.25]{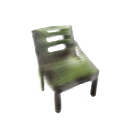}
                \end{minipage}
            \end{minipage}
            \hspace{-3mm}
            \begin{minipage}{0.20\textwidth}
                \centering
                \subcaption*{GT}
                \begin{minipage}{0.5\textwidth}
                    \includegraphics[scale=0.25]{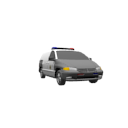} \\
                    \includegraphics[scale=0.25]{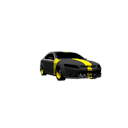} \\
                    \includegraphics[scale=0.25]{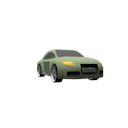} \\
                    \includegraphics[scale=0.25]{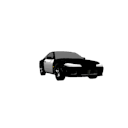} \\
                    \includegraphics[scale=0.25]{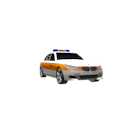} \\
                    \includegraphics[scale=0.25]{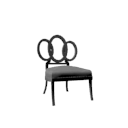} \\
                    \includegraphics[scale=0.25]{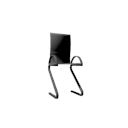} \\
                    \includegraphics[scale=0.25]{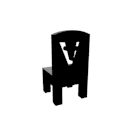} \\
                    \includegraphics[scale=0.25]{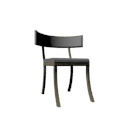} \\
                    \includegraphics[scale=0.25]{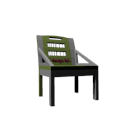}
                \end{minipage}
                \hspace{-3mm}
                \begin{minipage}{0.5\textwidth}
                    \includegraphics[scale=0.25]{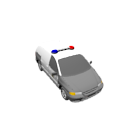} \\
                    \includegraphics[scale=0.25]{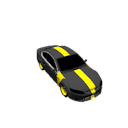} \\
                    \includegraphics[scale=0.25]{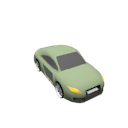} \\
                    \includegraphics[scale=0.25]{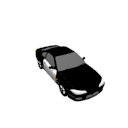} \\
                    \includegraphics[scale=0.25]{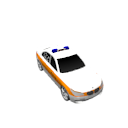} \\
                    \includegraphics[scale=0.25]{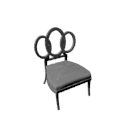} \\
                    \includegraphics[scale=0.25]{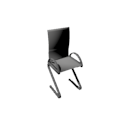} \\
                    \includegraphics[scale=0.25]{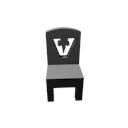} \\
                    \includegraphics[scale=0.25]{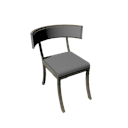} \\
                    \includegraphics[scale=0.25]{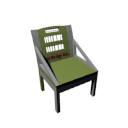}
                \end{minipage}
            \end{minipage}
        \end{minipage}
    \end{minipage}
    \caption{Additional qualitative evaluation of different configurations of our method on the ShapeNet-SRN~\cite{chang2015shapenet,sitzmann2019scene} dataset. (a): a minimalist version of our method only including the image encoder network and the hypernetwork, without taking pixel-aligned features and symmetric features as input to the neural radiance field. (b): adding pixel-aligned image features, compared to (a). (c): adding pixel-aligned and symmetric image features, compared to (a). (d): removing the hypernetwork, compared to (c).}
    \label{fig:ablation-more}
\end{figure*}

We also show more qualitative evaluation of different configurations of our method on the ShapeNet-SRN~\cite{chang2015shapenet,sitzmann2019scene} dataset in Fig.~\ref{fig:ablation-more}. The baseline model (a) tends to render smoothly. Simply using pixel-aligned image features (b) still fails to fully understand 3D structure. In contrast, our full model (c) reproduces photo-realistic details from most viewpoints. The rendering quality of (d) deteriorates as the hypernetwork is not adopted.
We have to emphasize that, \textit{only including both the symmetry priors and the hypernetwork can accurately recovers the geometry information and texture details despite the occlusions.}

\subsection{Network Architecture of SymmNeRF}
\begin{figure*}
    \centering
    \includegraphics[scale=0.63]{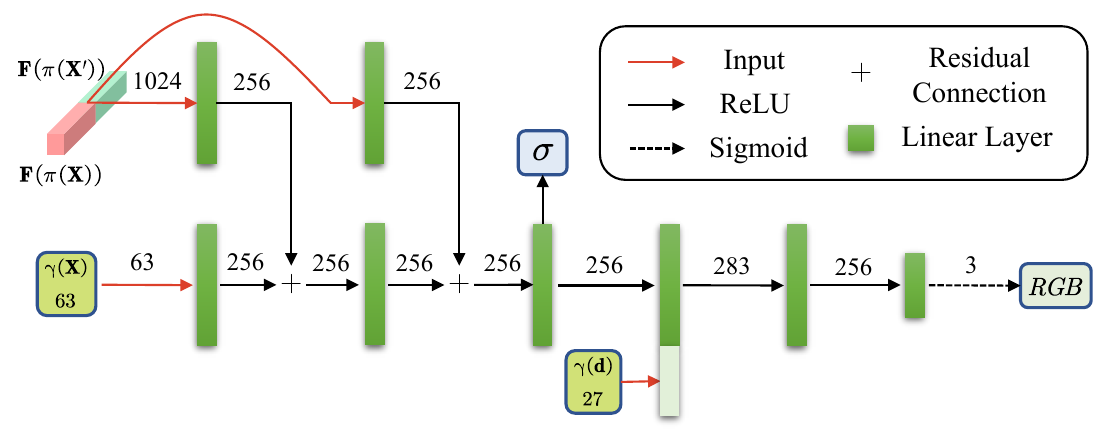}
    \caption{SymmNeRF architecture. $\mathbf{X}^{\prime} \in \mathbb{R}^{3}$ is the corresponding symmetric 3D point of $\mathbf{X}$, $\pi$ denotes the process of projecting the 3D point onto the image plane using known intrinsics, $\mathbf{F}$ is the feature volume extracted by the image encoder network $f$, and $\gamma_{\mathbf{X}}(\cdot)$ and $\gamma_{\mathbf{d}}(\cdot)$ are positional encoding functions for spatial locations and viewing directions, respectively.}
    \label{fig:architecture}
\end{figure*}

Here we visualize the network architecture of SymmNeRF in Fig.~\ref{fig:architecture}.

\begin{figure}[t]
    \centering
    \includegraphics[scale=0.2]{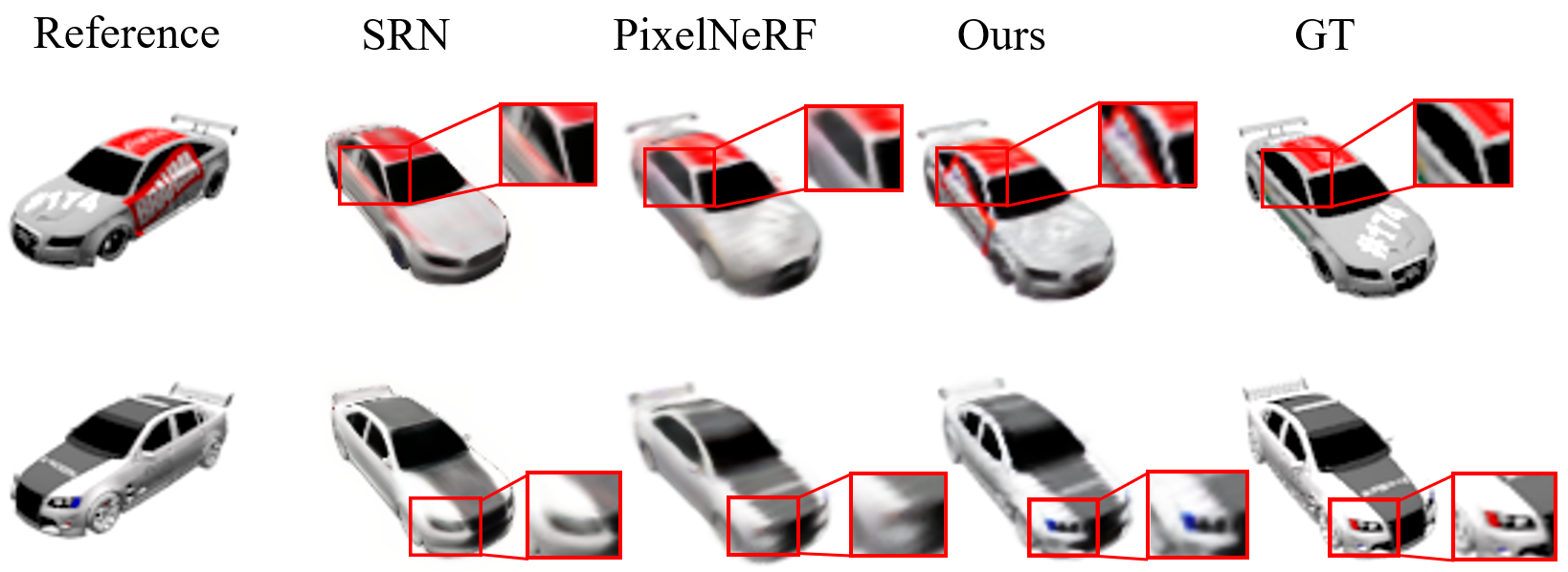}
    \caption{Failure cases. When the object is not perfectly symmetric and the reference view is not informative enough, symmetry priors may sometimes lead to erroneous rendering. }
    \vspace{-5mm}
    \label{fig:failure}
\end{figure}

\subsection{Limitations and Failure Cases}
Although symmetry can benefit single-view view synthesis, our method still suffers from some limitations. One is that when the object is not perfectly symmetric, symmetry priors may sometimes lead to erroneous rendering, see Fig.~\ref{fig:failure}. This depends on how informative the reference view is. Another one is that the model trained on single-object scenes may not handle multiple-object scenes, because multiple objects, as a whole, are usually asymmetric.

\end{document}